# Moral consensus and divergence in partisan language use

Nakwon Rim[a,b]*, Marc G. Berman[a,c], Yuan Chang Leong[a,c]*

[a] Department of Psychology, University of Chicago

[b] Knowledge Lab, University of Chicago

[c] The Neuroscience Institute, University of Chicago

**Abstract**

Polarization has increased substantially in political discourse, contributing to a widening partisan divide. In this paper, we analyzed large-scale, real-world language use in Reddit communities (294,476,146 comments) and in news outlets (6,749,781 articles) to uncover psychological dimensions along which partisan language is divided. Using word embedding models that captured semantic associations based on co-occurrences of words in vast textual corpora, we identified patterns of affective polarization present in natural political discourse. We then probed the semantic associations of words related to seven political topics (e.g., abortion, immigration) along the dimensions of morality (moral-to-immoral), threat (threatening-to-safe), and valence (pleasant-to-unpleasant). Across both Reddit communities and news outlets, we identified a small but systematic divergence in the moral associations of words between text sources with different partisan leanings. Moral associations of words were highly correlated between conservative and liberal text sources (average $\rho$ = 0.96), but the differences remained reliable to enable us to distinguish text sources along partisan lines with above 85% classification accuracy. These findings underscore that despite a shared moral understanding across the political spectrum, there are consistent differences that shape partisan language and potentially exacerbate political polarization. Our results, drawn from both informal interactions on social media and curated narratives in news outlets, indicate that these trends are widespread. Leveraging advanced computational techniques, this research offers a fresh perspective that complements traditional methods in political attitudes.

* Correspondence to: Nakwon Rim, nwrim@uchicago.edu, and Yuan Chang Leong, ycleong@uchicago.edu

**Introduction**

Political polarization is rapidly emerging as a profound threat in many societies worldwide, with the United States being a prime example (1–3). Although the extent of polarization within the broader public remains a topic of debate (4–6), studies have consistently found rising animosity and hostility between supporters of different political parties or ideological groups, a phenomenon known as "affective polarization" (7). Additionally, members within each group are increasingly converging on similar positions on key issues, leading to stronger internal alignment, a phenomenon referred to as "partisan sorting" (5,8,9). Affective polarization and partisan sorting exacerbate societal divides, hampering bipartisan efforts. The negative consequences of these trends are evident in the increased political gridlock and escalating social tensions (1,10). Identifying the key divisions among partisans will be crucial in understanding and mitigating the effects of political polarization.

Language is a revealing lens through which to view the partisan divide. In recent years, there has been a marked shift in the way people communicate about politics, with more negative, dismissive, or dehumanizing language becoming commonplace (11–16). Social media platforms, driven by algorithms that favor sensationalism, reinforce this divisive language and amplify extreme views (17) (but see (18)). Meanwhile, mainstream media, influenced by their ideological leanings, may also propagate polarized language and perpetuate divisions (19). While much of the previous research has centered around the tone of political discourse and its impact on political engagement or public opinion, less is known about how language in social and mainstream media reflects the beliefs and values of different political groups. In particular, the choice of words and phrases can reveal biases, beliefs, and values held by different political groups (20). For instance, a liberal group might frequently use terms such as "social justice", "equity", or "climate change", while a conservative group might emphasize "limited government", "family values", or "individual rights". The frequency, context, and sentiment associated with such terms can provide insights into the emotional resonance and importance these issues hold for each group.

Recent advances in natural language processing (see (21)) offer exciting opportunities to analyze textual data from social and mainstream media at large scales. One promising approach centers on the use of word embedding models, which map individual words in a corpus onto a high-dimensional vector. The location and distance between vectors are known to reflect semantic relationships (22). For example, the vectors of "king" and "queen" would likely be close to one another in the embedding space, denoting their related meanings; while both would likely be distant from unrelated words (e.g., "apple"). Recent studies have found that word embedding models often mirror deeply entrenched societal biases, such as those related to gender, race, and other cultural dimensions (23–25). For instance, Caliskan and colleagues (24) showed that, relative to male names, female names were more closely associated with family-related words than career-related words in a model trained on texts from the internet, echoing prevailing gender stereotypes. Word embedding models, trained on diverse corpora, can thus offer a unique window into various societal biases and associations embedded within a dataset. Building on this approach, recent studies have used word embedding models to quantitatively study differences in stereotypes and biases across time (26), languages (27), and age groups (28) by training models on relevant text sources.

In the current study, we capitalized on the unique opportunities afforded by word embedding models and the growing prevalence of political discourse in digital spaces. Specifically, we employed word embedding models trained on two sets of large-scale, real-life partisan corpora to shed light on how semantic associations of words diverged between conservatives and liberals. The first set of models was trained on user comments on Reddit, one of the world's largest social media platforms. These comments were sourced from 69 online discussion communities with a political focus and a strong partisan bias, providing us with a rich dataset of authentic discourse that arises informally in highly polarized environments (294,476,146 comments in total). The second set was trained on news and opinion articles from 38 news outlets with either a conservative or a liberal leaning (29) (6,749,781 articles in total). Together, the two sets of models allowed us to examine political language in both user-generated conversations and curated editorial content, enabling us to assess whether the findings are generalizable across digital platforms.

Our approach utilized the increasingly popular method of characterizing semantic dimensions, such as "large-small", "rich-poor", or "masculine-feminine", within an embedding's high dimensional vector space (30–35). In our study, we focused on the dimensions of *morality* (moral-to-immoral) (20,36,37) and *threat* (threat-to-safety) (38,39), both of which have been posited to be central to ideological leanings and how different political groups react to societal issues. Additionally, we explored the dimension of *valence* (pleasant-to-unpleasant) (40,41) to assess how affective associations differ along partisan lines.

First, we tested if we could see evidence of intergroup bias – that words related to the group that shared the political beliefs (in-group) will have more positive associations compared to words related to the group with opposing beliefs (out-group). For each corpus (Reddit and news), we projected words related to Democrats and Republicans onto the dimensions of morality, threat, and valence and examined how the relative position of in-group and out-group words differed by the political bias of the corpora. That is, for a liberal-leaning model, would words such as "democrats" and "liberals" be associated with greater morality, less threat, and positive valence relative to words such as "republicans", and "conservatives"? Thus, this analysis provided a proxy measure of beliefs and attitudes towards the political in-group and political out-group and allowed us to test the extent to which semantic associations in partisan corpora mirror real-world intergroup bias and affective polarization.

In addition to examining political in-group and political out-group terms, we assessed how semantic associations of words related to seven politically polarized topics (e.g., abortion, immigration, guns) differed between conservative and liberal corpora along each dimension (i.e., morality, threat, and valence). For each dimension, we tested for "semantic divergence" - evidence that word associations were more similar between corpora of the same ideological leaning than between corpora of different leanings. Such semantic divergence would be consistent with theoretical and empirical work suggesting that the same words and phrases are understood differently by individuals who do not share the same political ideology (20,42,43). By examining the semantic divergence of political words along morality, threat, and valence dimensions, our work uncovers the specific axes along which the meaning of politically charged terms and topics diverge between conservatives and liberals.

Altogether, our study utilizes quantitative text analysis on large-scale digital data from authentic discourse on social and mainstream media to investigate political attitudes. Using this approach, we document diverging semantic associations between conservative and liberal viewpoints, as well as the dimensions along which these divergences occur. The findings thus contribute to an empirical basis for understanding how the meaning of words might differ between individuals with different political ideologies. Our method of analyzing natural text generated by individuals in real-world contexts complements experimental and survey-based approaches to studying political attitudes. By combining these tools, future work can take a multifaceted approach to provide a more holistic understanding of political discourse and how it shapes inter-group attitudes, public opinion, and policy preferences.

## Results

**Overview of models and corpora.** We used *word2vec* models (22), a commonly used word embedding model, as the main tool of analysis. The first set of models ("Reddit models") was trained on user comments on Reddit between 2016-2021. We restricted our analyses to communities ("subreddits") that were political in nature and exhibited a strong partisan bias (see **Materials and Methods**). Separate models were trained on comments from each subreddit. For each subreddit, we obtained the partisan bias score computed by Waller and Anderson (44). We relied on this measure of partisan bias because it was computed solely from user/subreddit interaction data (i.e., in what subreddits users commented in) and did not consider the content of the posts. As such, our analyses of the semantic content of the comments would not be confounded by the computation of the partisan bias score. Following the approach by the authors (44), we defined conservative subreddits as those with a partisan bias score of 1 standard deviation higher than the mean (e.g., r/Conservative, r/TheDonald), and liberal subreddits as those with a partisan bias score of 1 standard deviation lower than the mean (e.g., r/democrats, r/hillaryclinton) (**Figure 1**). This yielded 28 conservative subreddits and 41 liberal subreddits (Table **S1**).

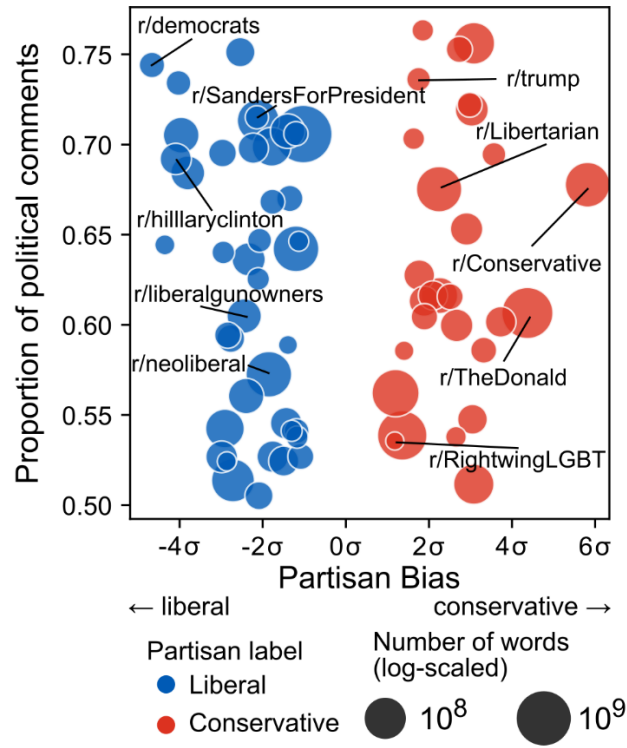

**Figure 1**. Word embedding models were trained on sixty-nine subreddit communities. Blue and red dots denote the liberal and conservative subreddits, respectively. The x-axis denotes the partisan bias scores calculated by Waller and Anderson (44), where positive numbers indicate conservative leanings while negative numbers indicate liberal leanings. Subreddits with a partisan bias score 1 standard deviation above the mean were categorized as liberal subreddits, while those with a partisan bias score 1 standard deviation below the mean were categorized as conservative. The y-axis denotes the estimated proportion of political comments calculated by Rajadesingan et al. (45).

The second set of models ("News models") were trained on news and opinion articles from 38 print media outlets between 2015-2019 by Rozado and al-Gharbi (29). The outlets were grouped into conservative and liberal outlets based on the Media Bias rating provided by Allsides.com (46). Allsides.com takes a multi-method approach to estimate political bias (e.g., editorial reviews, blind bias surveys, community feedback) and is commonly used by researchers studying political bias in news outlets (29,47,48). This yielded 15 conservative outlets and 23 liberal outlets (Table **S2**). Analogous to the Reddit models, a separate model was trained on articles from each outlet.

**Word projections along dimension vectors reflect semantic associations.** To study people's opinions on an issue, researchers often use behavioral surveys where individuals rate their views along predefined dimensions. For example, participants may be asked to rate how moral they think "abortion" is on a 1-7 scale from "very immoral" to "very moral." The average ratings can then be compared between political groups to study how attitudes differ by group. Here, we take an analogous approach using word embedding models to study how individual words are situated differently along predefined dimensions between political groups in real-life language use.

Word embedding models represent individual words as vectors in a high-dimensional vector space. When there is no risk of ambiguity, we refer to the vector corresponding to a word simply by the word for brevity (e.g., we refer to the word vector for the word "abortion" as "abortion"). To quantify the relationship of word vectors with predefined dimensions, we utilized the method of projecting words onto dimension vectors that reflect semantic dimensions of interest (30–34). We constructed dimension vectors by selecting words that define the two ends of a dimension (see Table **S3** for the list of words) and averaging the corresponding word vectors to define the respective "poles" of the dimension. For example, we constructed the morality dimension vector by averaging the word vectors for words such as "moral", "ethical", and "virtuous" to define the "moral" pole, and the word vectors for words such as "immoral", "unethical", and "wicked" to define the "immoral" pole. We then determined the positions of words along these dimensions (e.g., where does "abortion" lie on the "moral-to-immoral" dimension?). In mathematical terms, this involves calculating the scalar projections of the word vectors onto the dimension vector (see **Materials and Methods**). For simplicity, we describe the relative scalar projections of word vectors as relative positions to a pole. For example, if "kindness" has a higher scalar projection than "cruelty" when projected onto the morality dimension vector, we will describe it as "'kindness' was closer to the moral pole than 'cruelty.'" This then served as our framework for probing semantic relationships in a corpus.

We constructed three dimension vectors for each model: 1) *morality* (moral-to-immoral), 2) *threat* (threat-to-safety), and 3) *valence* (pleasant-to-unpleasant). We first assessed if the scalar projections of words along these dimensions aligned with human judgments. As an illustration, we would expect "kindness" to be closer to the moral pole than "cruelty" along the morality dimension, reflecting its stronger moral association. To formally assess this alignment, we projected words from external dictionaries that have been constructed to match human judgments. Specifically, we sourced moral and immoral words from the moral foundation dictionary 2.0 (49), threat words from Choi and colleagues' dictionary (50), and words with valence ratings from the NRC-VAD Lexicon (51). After projecting these words onto their respective dimension vectors, we statistically evaluated their alignment with dictionary-based judgments. For instance, we examined if moral words were positioned closer to the moral pole relative to immoral words (refer to **Materials and Methods**). The alignment was statistically significant in each individual corpus across all three dimensions in both the Reddit models and News models (all $p$s < .001, **Figures S1-S7**). Overall, these tests showed that the scalar projections of word vectors onto dimension vectors capture the expected semantic association of words within our models. This alignment provides validation for our approach of using these scalar projections to further investigate the semantic dimensions of target words in our subsequent analyses.

**Word associations in partisan corpora mirror real-world intergroup bias.** Partisans are known to favor in-group members (i.e., people who share their political beliefs) and dislike out-group members (i.e., people who are in opposing political groups), consistent with prior work on intergroup biases (52). Here, we test if a similar intergroup bias would be reflected in semantic associations in social and mainstream media platforms. To that end, we assessed if words related to Republicans (e.g., "republicans", "conservatives") and Democrats (e.g., "democrats", "progressives"; see Table **S4**) are positioned differently along the valence dimension by partisan bias of the corpora using permutation tests (see **Materials and Methods**).

In the Reddit models, the extent to which Republican-related words were closer to the "pleasant" pole relative to Democrat-related words was greater for models trained on conservative-leaning corpus than those trained on liberal-leaning corpus ($p < .001$, **Figure 2**). In other words, Republican-related words were closer to the "pleasant" pole than Democrat-related words in models trained on corpora that leaned conservative and vice versa, indicating an intergroup valence bias in user comments on Reddit.

Rozado and al-Gharbi (29), the authors who had trained the News models, had previously shown a similar result in those models - for conservative-leaning outlets, conservative-associated terms were more strongly associated with positive sentiment words than liberal-associated terms, with the opposite relationship observed for liberal-leaning outlets. We conceptually replicated their results using our approach described above ($p < .001$, **Figure S8**; see **Supporting Information** for the difference between the approaches).

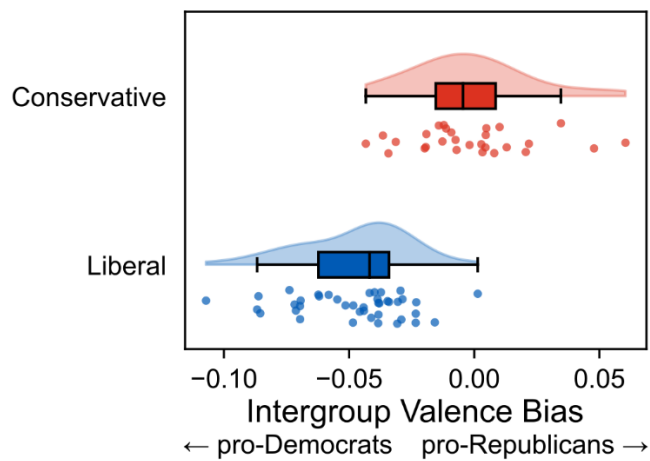

**Figure 2**. Valence association of words related to Democrats and Republicans differed by partisan bias of the subreddit the models were trained on. Intergroup valence bias refers to the degree to which Republican-related words were closer to the "pleasant" pole than Democrat-related words. The models were labeled using partisan bias scores calculated from user-interaction data by Waller and Anderson (44). Each point denotes data for one subreddit.

Similar results were shown with the morality ($p < .001$ for both Reddit and news; **Figure S9a,b**) and threat ($p < .001$ for both Reddit and news; **Figure S9c,d**) dimensions such that in-group-related words were associated with more morality and less threat compared to out-group-related words. Overall, these results indicate that in-group-related words would be positioned more positively on the valence, morality, and threat dimensions compared to out-group-related words in models trained on partisan text sources.

**Semantic divergence of issue words between partisan corpora.** Liberals and conservatives are known to disagree on many key political issues (e.g., abortion (53)). Our next analyses focused on whether and how this disagreement is reflected in partisan texts. Specifically, we investigated whether semantic associations of terms related to polarizing issues diverged between liberal and conservative corpora. We defined "semantic divergence" as the extent to which word associations differed between liberal and conservative corpora along a semantic dimension. We first chose words (including bigrams such as "gun_control") related to 7 politically polarized topics: abortion, constitution, guns, immigration, the LGBTQ+ community, police and criminals, and religion (Table **S5**). The position of these words along the morality, threat, and valence dimensions were then computed to extract their semantic associations along these dimensions. As an illustration, **Figure 3a** shows the position of a subset of issue words along the morality dimension, contrasting scalar projections from a conservative subreddit and a liberal subreddit.

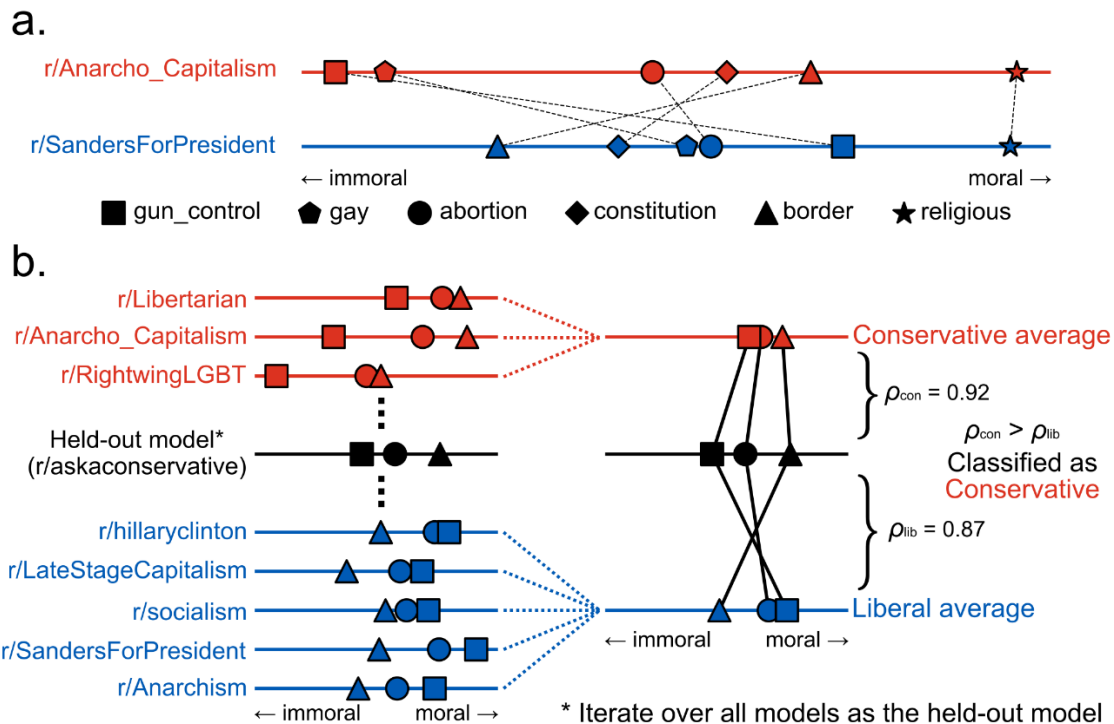

**Figure 3.** Assessing semantic divergence of issue words along dimensions. **a**. Semantic divergence along the morality dimension between a conservative (r/Anarcho_Capitalism) and a liberal (r/SandersForPresident) model for a subset of issue words. The positions of words reflect their scalar projections, with higher values (i.e., closer to "moral pole") on the right. The rank order of scalar projections of words is different between the two models. **b**. Schematic of classification analyses. One model (black; r/askaconservative) is chosen as the held-out model for this iteration. Of the remaining models, the scalar projections for all liberal models (blue models on the left) are averaged into a liberal average (the blue model on the right), and the scalar projections for all conservative models (red models on the left) are averaged into a conservative average (the red model in the right). The rank correlation (Spearman's $\rho$) between the held-out model's scalar projections and each average is calculated. The held-out model is classified as the group it is more correlated to. For example, the model shown here was classified as conservative. Note that this visualization shows the true scalar projection of a few selected words, but the analyses include a larger number of words.

We adopted a similarity-based classification approach (54) to assess semantic divergence. If semantic associations diverged between conservative and liberal corpora, we would expect the associations to be more similar between corpora with the same political leaning than between corpora of different leanings. A classifier would then be able to distinguish between conservative and liberal corpora based on whether the scalar projections along a dimension were more similar to that of the typical conservative or typical liberal corpora. Classification accuracy was computed following a leave-one-out-cross-validation framework, separately for each semantic dimension (**Figure 3b**). For each cross-validation fold, we held out the scalar projections of one model. Of the remaining models, we calculated the average scalar projections for each word separately for liberal and conservative models. These average scalar projections can be thought of as "template" scalar projections, reflecting the semantic associations along the dimension for the typical liberal or conservative model. We then calculated the rank correlation (Spearman's $\rho$) between the scalar projections of the held-out model and the liberal and conservative averages. If the correlation was higher with the conservative average, the held-out model was more similar to the typical conservative model and thus classified as a conservative model. In contrast, if the correlation was higher with the average liberal model, the held-out model was classified as a liberal model. This process was repeated, holding out a different model in each cross-validation fold, with the classification accuracy averaged across all folds. This analysis was performed separately for the Reddit models and the News models.

For the Reddit models, classification accuracy along the morality dimension showed the highest accuracy (88.41%, or 61 correct out of 69; **Figure 4a**). This was followed by valence (86.96%, or 60 correct out of 69) and threat dimensions (82.61% or 57 correct out of 69). We assessed statistical significance in two ways. First, we performed a permutation test by re-running the analyses 10,000 times after randomly shuffling the ideological label of the models. The empirical accuracy value was then compared to the null distribution generated by this procedure to calculate a p-value (yellow histograms in **Figure 4a**), which was then corrected for multiple comparisons across the three dimensions by controlling for the False Discovery Rate (FDR; (55)). All three dimensions demonstrated accuracy levels that were significantly above what we term "baseline accuracy" – the accuracy expected if ideological labels were assigned randomly (all FDR-$p$s < .001). This permutation test, however, did not speak to whether the theoretically derived dimensions of morality, threat, and valence were meaningfully better at discriminating between liberal and conservative corpora relative to arbitrary, non-specific dimensions. Would any dimension in the embedding space be able to distinguish between the two sets of corpora?

To address this question, we generated a second null distribution by repeating the analyses 10,000 times with "random" dimensions with randomly selected words forming the two poles (green histograms in **Figure 4a**). This null distribution provides a measure of classification accuracy that can be obtained from non-specific semantic associations along arbitrary dimensions (see **Supporting Information** for additional discussion on above-baseline accuracy for the random dimensions). Comparing the empirical results against this more stringent null distribution thus allowed us to assess if the morality, threat, and valence dimensions had additional discriminatory power above and beyond what could be achieved using arbitrary dimensions. The morality (FDR-$p$ = .023) and valence (FDR-$p$ = .028) dimensions performed better than the random dimensions, suggesting that the classification accuracy along these two dimensions did not merely reflect broad, nonspecific divergence between liberal and conservative corpora. In contrast, classification accuracy along the threat dimension was not significantly better than the random dimensions (FDR-$p$ = .107).

We also assessed classification accuracy along three semantically meaningful but theoretically irrelevant dimensions: size (big-to-small), speed (fast-to-slow) and wetness (wet-to-dry) (see Table **S6**). While these dimensions performed better than baseline accuracy (size: 75.36% or 52 out of 69; speed: 72.46% or 50 out of 69; wetness 81.16% or 56 out of 69; all uncorrected $p$s < .001), they did not outperform random dimensions (size: uncorrected $p$ = .496; speed: uncorrected $p$ = .676; wetness: uncorrected $p$ = .155). These results provide further evidence that the classification results observed along the morality and valence dimensions did not merely reflect non-specific divergence between liberal and conservative corpora.

For the News models, all three dimensions similarly performed better than baseline (**Figure 4b**), with the morality dimensions again showing the highest accuracy (89.47% or 34 correct out of 38, FDR-$p$ < .001), followed by threat (73.68% or 28 out of 38, FDR-$p$ = .012) and valence (71.05% or 27 out of 38, FDR-$p$ = .022). However, classification accuracy was significantly better than random dimensions only for the morality dimension (FDR-$p$ = .033) and not the other two dimensions (both FDR-$p$ = .503). Of the three irrelevant dimensions (i.e., size, speed, and wetness), only classification along the speed dimension significantly outperformed baseline accuracy on the News models (size: 65.79% or 25 out of 38, uncorrected $p$ = .077; speed: 76.32% or 29 out of 38, uncorrected $p$ = 0.003; wetness: 63.16% or 24 out of 38, uncorrected $p$ = .138). Classification along all three irrelevant dimensions was not significantly better than random dimensions (size: uncorrected $p$ = .722; speed: uncorrected $p$ = .273, wetness: uncorrected $p$ = .811). Thus, of all dimensions tested, morality was the only dimension along which semantic divergence is greater than random dimensions across both Reddit and News models. Together, these results indicate that the divergence in moral associations of words related to political topics was sufficiently reliable to distinguish corpora along partisan lines.

In the above analysis, we focused on a selected set of issue words to specifically probe the semantic divergence of terms pertaining to political issues. In doing so, we ensured that the classification results would not be driven by non-issue related terms, including those that referred to in-group or out-group members. Nevertheless, as a robustness test of our findings, we expanded the classification analyses to include all common words shared between the models: 14,607 words in the Reddit models and 17,456 words in the News models. These analyses yielded the same pattern of results. First, classification along all dimensions showed higher than baseline accuracy across both Reddit and News models (all FDR-$p$ < .001). Second, for the Reddit models, classification along the morality (91.30%, FDR- $p$ = .007) and valence (Reddit: 94.20%, FDR- $p$ = .002) dimensions performed better than random dimensions, while threat did not (81.16%, FDR-$p$ = .383). Lastly, in the News models, classification accuracy was significantly better than random dimensions only along the morality dimension (89.47%, FDR-$p$ = .030) and not the valence (84.21%, FDR-$p$ = .096) and threat dimensions (81.58%, FDR-$p$ = .133). As such, our findings were consistent regardless of whether we used a focused set of issue words or a broader selection of common words. We note that these results do not necessarily mean that partisans diverged in "non-political" words. As the common words included political words (e.g., issues words, in-group and out-group words), they could still drive the classification results. Given the difficulty in exhaustively excluding all political words, we did not attempt to further segregate the common words into "political" and "non-political" words.

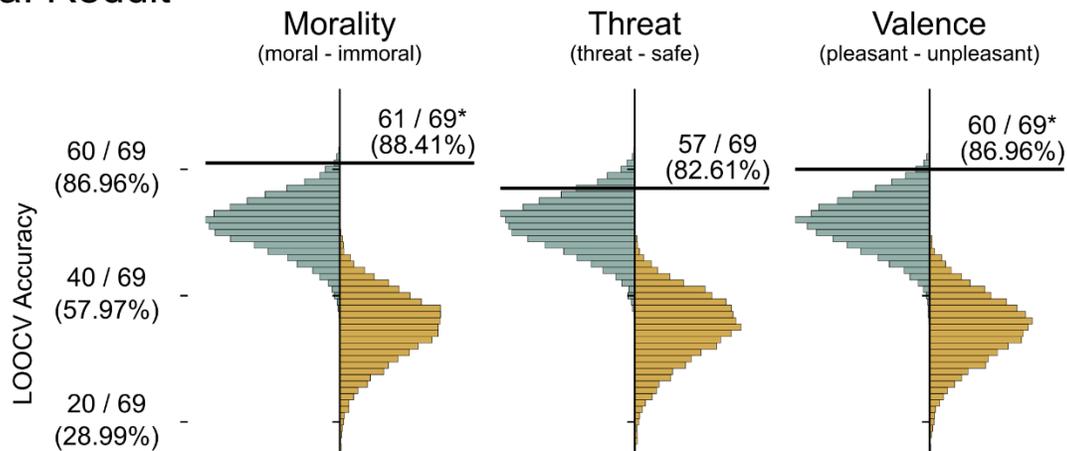

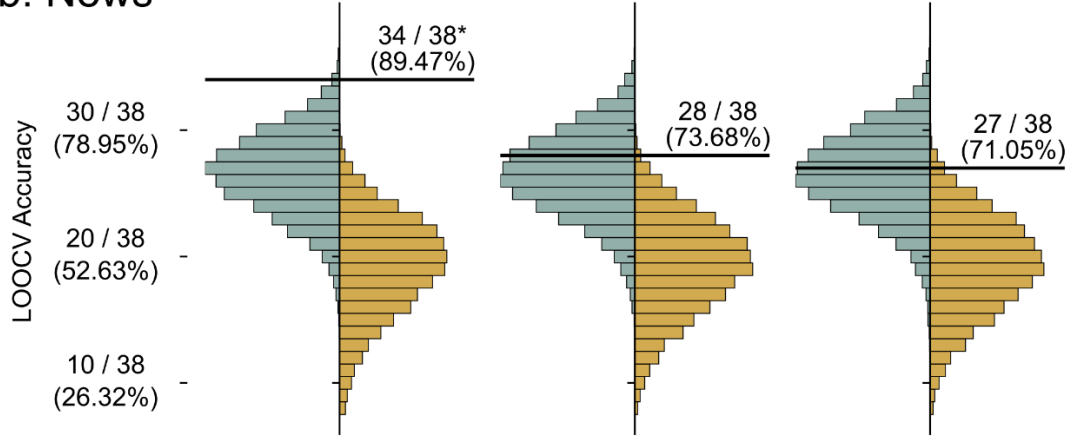

**Figure 4**. Leave-one-out-cross-validation classification results along semantic dimensions. **a**. Reddit models. The black line represents empirical accuracy. Yellow histograms (right of axis) show baseline null distributions from permuting partisan labels; green histograms (left of axis) depict null distributions from random dimensions with randomly selected word poles. Both histograms are based on 10,000 samples. **b**. News models. The visualization and elements are consistent with a. *: FDR-$p$ <.05.

**Positions of words between average conservative and liberal models were highly correlated.** The classification results indicated the semantic divergence between partisan corpora was sufficiently reliable such that it was possible to distinguish between liberal and conservative corpora with a high degree of accuracy, especially along the morality dimension. However, the results do not speak to the magnitude of the divergence. On the one hand, semantic divergence could be large and reliable, which would suggest that liberals and conservatives have vastly different semantic associations with issue-related words. This would paint a picture of two distinct ideological landscapes, where the way in which key issues are understood and conceptualized varies greatly between the two groups. On the other hand, the semantic divergence could be small but reliable.

In this scenario, while liberals and conservatives might have different associations with certain issue words, these differences may be subtle, where both groups largely share a common understanding, with small but reliable differences.

To address this question, we calculated the rank correlation between the scalar projections of issue words of the average liberal and average conservative model across each dimension. This correlation measured the degree to which the positions of issue words were, on average, similar between conservative and liberal models. The scalar projections of issue words were highly correlated between liberal and conservative models across all three dimensions in both Reddit (**Figure 5a**; morality: Spearman's $\rho$ = 0.97; threat: Spearman's $\rho$ = 0.92; valence: Spearman's $\rho$ = 0.94) and News models (**Figure 5b**; morality: Spearman's $\rho$ = 0.95; threat: Spearman's $\rho$ = 0.95; valence: Spearman's $\rho$ = 0.95). This high correlation indicates that while there are systematic differences that enable classification between partisans, the overall semantic landscape for both liberal and conservative models is remarkably similar, suggesting both political groups operate within a shared semantic space, with distinctions lying more in nuanced variations rather than in stark opposition.

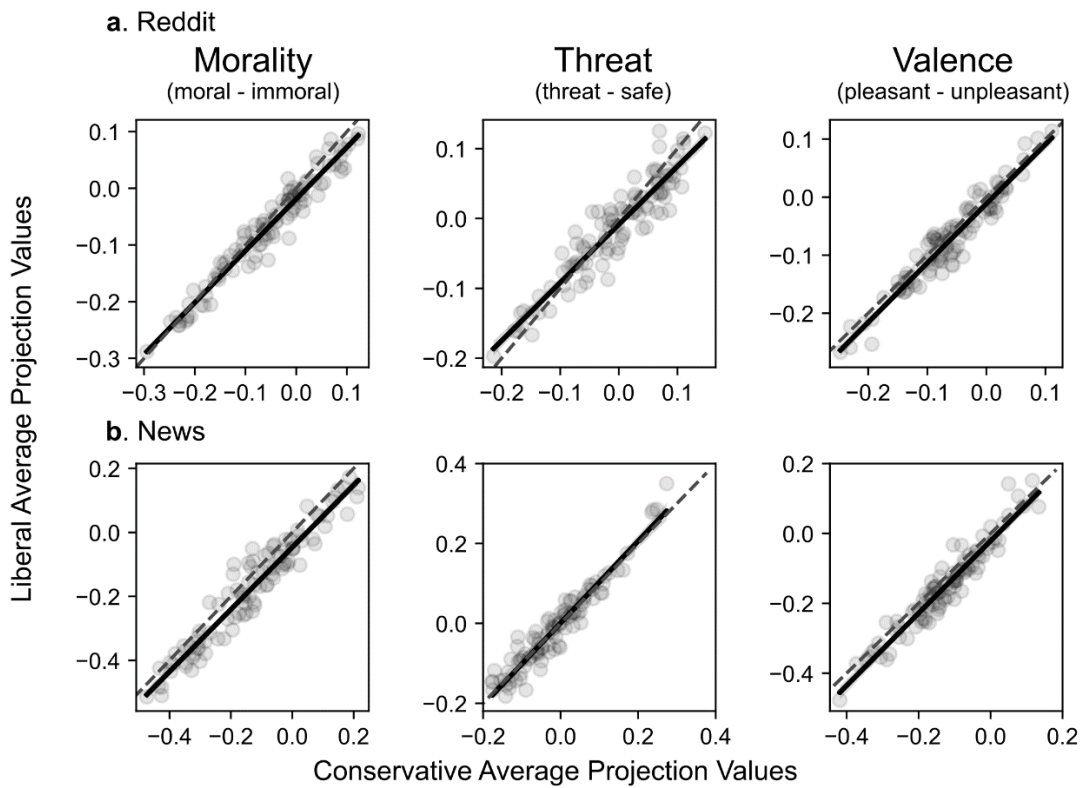

**Figure 5**. **a**. The correlation between the liberal and conservative average projection vectors is visualized for the Reddit models for all issue words. The columns correspond to results from morality, threat, and valence dimensions, respectively. The dots denote the average scalar projections for the word vectors, the solid lines are the best-fit lines using linear regression, and the dotted lines are the identity lines. **b**. The correlation between the liberal and conservative average projection vectors is visualized for the news models. The column structure and visual elements are the same as in **a**.

**Discussion**

In this study, we utilized word embedding models trained on two sets of large-scale corpora of real-world political discourse to quantitatively examine how semantic associations of words diverged between conservative and liberal perspectives. Across both platform types, we found that words related to the opposing political group were more strongly associated with negative valence, immorality, and greater threat. This result highlights how the phenomenon of affective polarization, characterized by negative perceptions of members of the political outgroup, manifests in real-world political discourse. Furthermore, we quantified the semantic associations of terms related to political topics along the psychological dimensions of morality, threat and valence. For each dimension, we tested for semantic divergence, that is, evidence that semantic associations of political words diverged between corpora of different leanings. Of the three dimensions examined, morality was the only dimension for which semantic divergence was greater than what was observed for arbitrary, non-specific dimensions across both Reddit comments and news articles. The pattern of semantic divergence along the morality dimension was sufficiently reliable such that a classifier could determine the ideological leaning of corpora with high accuracy, suggesting that moral connotations associated with political topics may be distinctly interpreted by conservative and liberal viewpoints. However, moral associations of political words remained highly correlated between corpora with different ideological leanings, indicating that the magnitude of semantic divergence was small. These results underscore a shared moral understanding across the political spectrum, with small but consistent differences that may nevertheless hinder dialogue across political groups.

Affective polarization, characterized by negative perceptions of members of the political outgroup, fosters intergroup hostility (7). Evidence of affective polarization has largely depended on explicitly asking participants to state their perceptions of and feelings toward in-group or out-group members. These measurements, while invaluable, are influenced by the specific question asked, response biases, and the overt nature of direct questioning (56,57). Our findings examine affective polarization in the context of natural political discourse, a more implicit and pervasive medium through which individuals express and form attitudes. We found that user comments on Reddit, a platform where users engage in spontaneous and informal discussions, exhibited intergroup valence biases, highlighting the extensive influence of affective polarization on individual perceptions and conversations. This work extends results from Rozado and al-Gharbi (29), who found evidence that language used by mainstream media sources show patterns of affective polarization in their content. Given the vast reach of social media and mainstream news, the presence of these biases in such platforms could further amplify affective polarization, solidifying divisions among users and readers (58,59). By offering a quantitative measure of affective polarization that tracks with partisan bias, our approach can help researchers identify online communities and news sources with heightened affective polarization, which can inform policy recommendations or platform adjustments.

Semantic divergence, as we have defined in the current work, is grounded in the idea that the same words or terms can carry different connotations for conservatives or liberals (20,42,43). For example, "taxation" may evoke notions of excessive government intervention and an unfair infringement of individual autonomy for conservatives, while for liberals, it might symbolize a communal responsibility towards the provision of public goods. This difference may be challenging to probe using methods such as questionnaires and interviews, as people are not always able to provide accurate, explicit reports of their mental representations (60,61). Consequently, researchers have turned to creative and indirect means to uncover subtle differences in the meaning of words. One approach to probe semantic representations is to examine their relationship with other words (62,63). A recent study had conservative and liberal participants each place 30 political words in a two-dimensional space such that words that were more similar in meaning were placed closer together (64). The spatial configuration of words, which reflected each participant's belief about what words were associated with one another, was more distinct between participants who were further apart in political ideology, suggestive of divergent semantic representations. In the same study, the authors also used functional magnetic resonance imaging to measure participants' neural patterns while they viewed the same 30 words. Neural patterns, which have been previously shown to encode the semantic information of words (65,66), were more distinct between participants who were further apart in political ideology, providing converging evidence that semantic divergence of political words between individuals with different political beliefs.

Our work presents an alternative and complementary methodology to identify semantic divergence between political partisans. By analyzing semantic relationships embedded in vast amounts of real-world textual data, we examined how semantic divergence takes shape in wider public discourse. In addition, our findings extend prior work by assessing the dimensions along which the semantic divergence occurs. Our finding that moral associations of political words consistently diverged between conservative and liberal corpora provides empirical support for theories suggesting that ideological differences are rooted in distinct moral worldviews that shape values and attitudes (20,36,67). Consistent with these theories, recent work found that politicians tend to emphasize different moral concerns in their rhetoric based on their party affiliations, possibly catering to the moral frameworks of their supporters (68,69). Here, we focus on language use in online communities and news sources, demonstrating that the same words, when used in different ideological contexts, can have distinct moral connotations. This divergence in moral associations may both reflect and amplify the differences in moral evaluations surrounding political issues.

While the divergence along the morality dimension was reliable, as shown by the higher classification accuracy across both Reddit and News datasets, the moral associations of words remained highly correlated between conservative and liberal corpora. In other words, even though there were systematic differences in moral associations between conservative and liberal corpora, the magnitude of this divergence was not overwhelmingly large. This small difference is consistent with findings that Democrats and Republicans may not be as polarized on issues as popular narratives or in-group perceptions often depict (8,70,71). Similar to past work suggesting small biases for the in-group might be sufficient to drive large-scale segregation (72), the small differences in moral connotations might be enough to drive large political divides. Nevertheless, our results suggest that despite the differences in perceptions, there is also a shared moral understanding that transcends partisan boundaries. Recognizing this common moral ground may offer avenues for bridging divides and fostering dialogue between political groups. For example, emphasizing different moral values associated with a political issue has been shown to increase the receptiveness of individuals to alternative positions on the issue (37,73). Overall, our results highlight the importance of taking into account both the magnitude and reliability of political differences in understanding the divide between political partisans.

A strength of our results is that we found semantic divergences along the morality dimension across two highly distinct platform types. While Reddit allows for organic, user-driven conversations often spurred by immediate events or personal sentiments, mainstream news outlets present curated content that reflects journalistic standards and editorial choices. The consistency of our findings across both platforms underscores the pervasiveness of these semantic biases. Certainly, the dynamics of Reddit and mainstream news outlets do not capture the entirety of online discourse, and there may be other platforms or mediums where these patterns differ or evolve in unique ways. Future research can examine other corpora, including different online forums, social media platforms, and offline dialogues, to determine whether the observed semantic divergence persists across various populations, languages, and sociopolitical contexts.

Our findings did not identify threat as a core dimension of semantic divergence between conservative and liberal corpora. Across both Reddit and News corpora, classification accuracy along the threat dimension was not significantly better than arbitrary, non-specific dimensions. It is worthwhile to note that these findings do not contradict prominent theories that relate threat perception to political ideology, as these theories primarily emphasize heightened threat sensitivity in conservatives relative to liberals and do not make strong claims about whether the two groups differ in what they perceive as threats (38,39). As the scalar projections of words are in arbitrary units, we do not have a measure of the "threat-level" of a word in absolute terms. Thus, our method is ill-suited to test the central claim of these theories that conservatives tend to respond more strongly to threats than liberals.

More broadly, our analyses are centered around identifying semantic dimensions differentiating partisan corpora in the aggregated pattern across multiple words. We would advise caution when interpreting the position of individual words along a dimension, as they do not necessarily reflect the attitude towards an issue or group of people denoted by the term. For example, the word "immigrants" was, on average, closer to the immoral pole than the moral pole for liberal models relative to conservative models. One interpretation of this finding could be that liberals perceive immigrants as more immoral than conservatives do, which would be inconsistent with other work indicating that liberals tend to have more favorable attitudes towards immigrants (74,75). An alternative explanation is that in liberal corpora, the word "immigrants" co-occurs with discussions of the immorality of immigration policies, such as family separations at the border or indefinite detention of asylum seekers. Thus, while our approach can be used to identify points of divergence, the use of convergent methods will be critical when validating and teasing apart specific interpretations.

Indeed, we see our approach as a complement, and not a replacement, to surveys, experiments, and qualitative studies. Our method enables researchers to cast a wide net over expansive corpora efficiently and at minimal cost. Once the models are trained, they can be queried to explore different questions without new rounds of data collection. This is particularly beneficial in generating new hypotheses, which can then be rigorously tested and validated using more targeted research tools, such as surveys or controlled experiments. Additionally, our approach holds promise for studying dynamic shifts in discourse over time (26). Other work has shown that the prevalence of moral language in partisan discourse increases when one's party is in a position of lower political power (76), though it remains unclear if and how moral connotations differ based on changing political landscapes. Future work can analyze corpora from varying time periods to investigate how semantic associations evolve with cultural and political shifts.

In the current work, we used static word embeddings where each word is assigned the same vector representation irrespective of its context in a sentence. Our choice of models was motivated by the lower computational cost and the availability of analytical tools that have been established to capture human-like biases in semantic associations (23,24,28). A fruitful direction for future research is the use of contextual embedding models, where different vector representations are assigned to words based on their context within a sentence or paragraph (e.g., (77)). These models offer the potential to provide richer representations of words, capturing complex contextual nuances in partisan discourse. We are excited by the rapid advancement in these models and believe that the development of analytical approaches tailored to these models will significantly enhance our ability to understand the intricacies of semantic divergence in partisan rhetoric.

To conclude, our work demonstrates the viability and usefulness of using language models to provide a scalable method for studying divergent partisan perceptions embedded in political discourse. Using our approach, we revealed patterns of affective polarization and moral semantic divergence across two distinct types of online text. Our findings support existing theories on partisan divides that propose morality as a central dimension to the differences between partisans, demonstrating that these moral differences manifest in natural political discourse. The small magnitude of differences, however, suggests underlying similarities in moral understanding that could potentially be leveraged to bridge divides. Future research can combine the capabilities of language models with surveys and experiments to foster a deeper understanding of partisan perceptions and inform efforts that facilitate constructive dialogue across partisan boundaries.

**Materials and Methods**

**The Reddit *word2vec* models.** The Reddit comment corpus from 2016 to 2021 was downloaded from the Pushshift Reddit dataset (78) (https://files.pushshift.io/reddit/comments/) on 2022/06/23. From this data set, we first collected a list of subreddits that were labeled as liberal or conservative (-leaning) by Waller and Anderson (44). Waller and Anderson calculated the partisan bias scores based on user/subreddit interaction data (i.e., in what subreddits users commented) without considering the semantic content of the comments. After calculating and validating the scores, Waller and Anderson defined conservative(-leaning) and liberal(-leaning) subreddits as those with a partisan bias score 1 standard deviation higher or lower than the mean, respectively.

In addition to the partisan bias, we restricted the analyses to subreddits that discuss political topics. To achieve this, we utilized results from Rajadesingan and colleagues (45), who estimated the proportion of political comments for each subreddit. This was done with a classifier trained to distinguish between political and non-political comments on Reddit. Utilizing their results, we restricted our analyses to subreddits where at least 50% of the comments were classified as political in nature.

We then collected all comments from subreddits that were ideologically biased and political in nature. We preprocessed the comments in multiple steps, including removing comments from bot accounts, removing non-alphanumeric characters, and detecting collocations (see **Supporting Information**). After preprocessing, we additionally filtered out subreddits that had less than 10,000,000 word tokens in total or less than 30,000 unique word tokens that appeared at least five times. This was to ensure that there was sufficient data to train the word embedding models. This resulted in 69 subreddits listed in Table **S1**, which were used to train the *word2vec* models. There were, on average, 136,792,029 word tokens from 4,267,770 comments per analyzed subreddit.

The *word2vec* models were trained using the Gensim (79) package in Python version 3.9.12. There are numerous hyperparameters that affect the resulting embeddings of a word2vec model. To ensure that our results were not unduly affected by the choice of hyperparameters, we trained 64 different *word2vec* models for each subreddit using a soft grid search across three hyperparameters (context window size, number of negative words sampled, and downsampling rate of frequent words; see **Supporting Information**). For each word token, the scalar projections of word vectors in 64 models were then averaged for the main analysis of the paper.

**The news *word2vec* models.** We used pre-trained models from (29) for the news corpora analysis, downloaded from https://zenodo.org/record/4797464#.Ys5UpYRBzIw. The authors collected news and opinion articles from news outlets' websites and public repositories. The articles were preprocessed using steps including removing HTML data, markup syntax, and removing non-alphanumeric characters. After preprocessing, the authors trained *word2vec* models on news and opinion articles from years 2015 to 2019, individually for each news source. Analogous to our Reddit models, the Gensim package in python was used to train the models. Out of the 47 models, we analyzed 38 models trained on a source which the authors classified as liberal(-leaning) or conservative(-leaning) based on the media bias rating from AllSides.com Media Bias Chart version 1.1 (46). Allsides.com classifies outlets as "Left", "Lean Left", "Moderate", "Lean Right", and "Right." We treated outlets rated as "Left" or "Lean Left" as liberal, and outlets rated as "Right" or "Lean Right" as conservatives. The 38 news sources are listed in Table **S2**. There were, on average, 177,625 articles per analyzed outlet. Please refer to the original paper (29) for additional details on the News models.

**The common vocabulary sets.** Since each model was trained on different corpora, each model had a different set of vocabulary (i.e., word tokens that appeared at least five times in the corpora). As we are comparing across these models, we utilized word tokens that were common across all models: 14,607 word tokens for Reddit models and 17,456 word tokens for News models.

**Building, projecting onto, and validating dimension vectors**. We use the methodology proposed by (30,32) to build dimension vectors and project word vectors onto them. We first defined two sets of word vectors with opposing meanings along each target dimension ($S_1, S_2$). For example, for the morality dimension, we selected words like "moral", "ethical", and "virtuous" to define a moral pole $S_1 = \{\overrightarrow{moral}, \overrightarrow{ethical}, \overrightarrow{virtuous} ...\}$; $\overrightarrow{W}$ denotes word vector for word $W$), and words like "immoral", "unethical", and "wicked" to define an immoral pole ($S_2 = \{\overrightarrow{immoral}, \overrightarrow{unethical}, \overrightarrow{wicked} ...\}$). The first author first identified all words in the common set that corresponded to each pole. The resulting sets were then discussed and revised by all three authors until all three agreed on the set. The final sets are listed in Table **S3**. This process was done before any computation using these words was done. On average, each pole was defined by 33.58 words.

The pole vectors ($\boldsymbol{p_1}, \boldsymbol{p_2}$) were defined as the average of the word vectors in the corresponding set:

$$\boldsymbol{p_i} = \frac{1}{|S_i|} \sum_{v \in S_i} \hat{v}$$

where $\hat{v}$ denotes the unit vector in the direction of $v$ (i.e., $\hat{v} = \frac{v}{|v|}$).

Using the pole vectors, the dimension vector $\boldsymbol{d}$ is defined as:

$$\boldsymbol{d} = \widehat{\boldsymbol{p_1}} - \widehat{\boldsymbol{p_2}}$$

Finally, we found the position of a word vector $w$ when projected onto the dimension vector $\boldsymbol{d}$ by computing the scalar projection, defined as

$$\boldsymbol{w_d} = \boldsymbol{w} \cdot \widehat{\boldsymbol{d}}$$

Note that since the Gensim implementation of *word2vec* normalizes the word vectors (i.e., $\boldsymbol{w} = \widehat{\boldsymbol{w}}$), this is equivalent to the cosine similarity between $\boldsymbol{w}$ and $\boldsymbol{d}$ ($\widehat{\boldsymbol{w}} \cdot \widehat{\boldsymbol{d}}$).

To validate that the scalar projections along the dimension vectors reflect expected semantic associations, we projected words from external dictionaries onto the dimensions. For the morality dimension, we projected virtue and vice words (which we called moral and immoral words in the results) from the moral foundation dictionary 2.0 (49) onto the moral dimension vectors. We then tested if the virtue words were closer to the "moral" pole (i.e., have higher scalar projections) compared to the vice words using two-sample *t*-tests. For the threat dimension, we projected threat-related words from the dictionary created by Choi and colleagues (50) onto the threat dimension vector. We then tested if the words were closer to the threat pole compared to the median word vector using one-sample *t*-tests. For the valence dimension, we projected words from NRC-VAD Lexicon (51) onto the valence dimension vector. We then calculated the correlation (Pearson's *r*) of the valence rating of the words and the scalar projections. All tests were done separately for each subreddit/outlet, and only words that appeared in the common vocabulary set were used. **Supporting Information** contains additional details on the validation process.

**Measuring Intergroup bias in the models**. To test the difference in the semantic association of in-group and out-group words, we first selected Republicans-related and Democrats-related words from the common vocabulary set, following the procedure described for selecting pole words above (Table **S4**; 13 Republican-related words; 22 Democrat-related words in total). We then projected the Republicans-related and Democrats-related words onto the valence dimension vector for each subreddit or outlet. The scalar projections were averaged separately for the Republicans-related words and the Democrats-related words.

The average values for the Democrats-related words were subtracted from those for the Republicans-related words. This difference in average scalar projections reflected the extent to which Republican-related words were closer to the pleasant pole than the unpleasant pole, relative to Democrat-related words. We refer to this as the intergroup valence bias of a given outlet or subreddit. We then tested if the intergroup valence bias was significantly different by the political bias of the corpora models were trained in, labeled by partisan bias score calculated by Waller and Anderson (44) for Reddit models, and media bias rating given by Allsides.com (46) for News models. This was done using a permutation test, where we compared the absolute difference in mean of the original grouping (liberal and conservative models) against the absolute difference in means of 10,000 randomly permuted grouping. Specifically, we calculated the *p*-value as

$$P = \frac{1 + \sum_{n \in N} I_{n \geq T}}{1 + |N|}$$

where $N$ is the null statistics (absolute difference in mean) from the sampled null distribution (permuted grouping) and

$$I_{n \geq T} := \begin{cases} 1 \text{ if } n \geq T \\ 0 \text{ if } n < T \end{cases}$$

where $T$ is the empirical statistics to be tested (absolute difference in mean from the original grouping).

In other words, the *p*-value is defined as (1 + number of null values that are the same or larger than the tested value) / 10,001. The process was repeated using the morality dimension vector and the threat dimension vector.

**Calculating classification accuracy and performing significance testing for semantic divergence analysis.** To assess the semantic divergence of issue words in the trained models, we first selected words (including bigrams) related to seven politically polarized topics: abortion, constitution, guns, immigration, the LGBTQ+ community, police and criminals, and religion (Table **S5**; average 14.29 words per topic). This was done by the procedure used to choose the pole words. We then calculated the scalar projection of the word vectors for the selected words onto the three dimensions (morality, threat, and valence) for each corpus.

We used a similarity-based classifier (54) to assess divergence in the scalar projections of the issue words. The leave-one-out-cross-validation framework was used to calculate the classification accuracy separately for the three dimensions. For each cross-validation fold, we held out the scalar projections of one model and calculated the average scalar projections for each word separately for the remaining liberal and conservative models. We then calculated the rank correlation (Spearman's $\rho$) between the scalar projections of the held-out model and the liberal and conservative averages. If the correlation was higher with the conservative average, the held-out model was classified as a conservative model, and vice versa. This process was repeated, holding out a different model in each fold. The classification accuracy was then averaged across all cross-validation folds. This analysis was performed separately for the Reddit and the News models.

We sampled from two null distributions to statistically test the classification accuracy. The first null distribution was sampled by permuting the label (liberal or conservative) of the subreddits/outlets and calculating the classification accuracy for each dimension 10,000 times. We refer to this as the "baseline accuracy." To sample the second null distribution, we first randomly sampled words without replacement from the common vocabulary set for each pole, matching the number of words used to construct the original pole vector. For example, when building the null distribution for the morality dimension in Reddit, we drew 27 words randomly for one pole and 31 words randomly for the other pole, corresponding to the number of words at the moral and immoral pole respectively. Then, we projected word vectors into this newly sampled random dimension to obtain the scalar projections and calculated the classification accuracy. The process was repeated 10,000 times for each dimension to sample the null distribution. This null distribution reflected the accuracy when classifying along arbitrary, non-specific dimensions.

For both null distributions, the *p*-value was calculated as the same fashion as the permutation test stated above. All *p*-values reported using this test are corrected for false discovery rate (FDR) using Benjamini-Hochberg procedure (55) with $\alpha = 0.05$. Finally, the whole procedure was repeated using all common words instead of issue words.

**Calculating correlations of scalar projections between average conservative and liberal models.** To investigate the alignment of scalar projections between average conservative and liberal models, we first averaged scalar projections of the issue-related words separately for all liberal and conservative models. We then calculated the rank correlation using Spearman's $\rho$ between liberal average and conservative average, separately for each dimension. This analysis was conducted separately for the Reddit models and the News models.

**Acknowledgments**

This research was supported by a seed grant from the University of Chicago Data and Democracy Initiative to YCL, a National Science Foundation Smart and Connected Communities grant (NSF S&CC) #1952050 from the Division of Computer and Network Systems (CNS) to MGB, and internal grants from the University of Chicago to MGB and YCL. This work was completed in part with resources provided by the University of Chicago's Research Computing Center.

We thank members of the Department of Psychology and Knowledge Lab at the University of Chicago, as well as Dr. James Evans, Dr. Joshua Conrad Jackson, Dr. Austin Kozlowski, Dr. Alex Shaw, Hayoung Song, and Dr. Robb Willer (names listed alphabetically), for helpful feedback on the project.

**Data and Code Availability**

The data and code to replicate the results will be available for the peer-reviewers during the review process and will be publicly released after formal publication.

**Supporting Information Text**

**Additional detail on the Reddit corpora and models**

For the Reddit corpora, all comments written by known bot accounts (1) were removed from the analysis. We also combed through randomly sampled comments to identify additional bot accounts, which were also removed from the analysis (listed below). We additionally removed comments that contained phrases that indicate moderating activities (e.g., notifying that the moderators removed a comment because it violated the subreddit's policy) or announced that the comment was made by a bot. The phrases used are listed below. After removing the comments in the previously mentioned criteria, we threw out word tokens that were determined as punctuation, number, or URL by SpaCy (https://spacy.io/)'s *en_core_web_lg* model. Finally, we removed any non-alphanumeric characters from the corpus.

After preprocessing the corpus, we trained the collocation detection model implemented in the Gensim package (2–4) on the corpus as a whole (i.e., one collocation detection model was trained across all 69 subreddit corpora). We applied this model to the corpus to merge tokens that the model identified as collocations (e.g., tokens "gun" and "control" were merged to "gun_control" when they appeared together).

Finally, 64 *word2vec* models were trained for each subreddit corpora. All models were trained using the skipgram algorithm with negative sampling. All models were set to have 300 dimensions (i.e., the size of each word vector was set to be 300), and tokens that appeared less than 5 times were ignored. The 64 models had different combinations of hyperparameter values in window size (5, 10, 15, 20), number of negative words sampled (5, 10, 15, 20), and downsampling rate (0.1, 0.01, 0.001, 0.0001). For all other hyperparameters, the default value in the Gensim implementation was used.

**List of additionally removed bot accounts**. *jobautomator, TheWallGrows, Decronym, WorldNewsMods, PoliticalHumorMods, TwitterToStreamable, TheWallGrowsTaller, RIPmod, CrookedPrisoner, revddit, worldpolitics, properu, tacostats, EmojiPoliceGetDown, twinkiac, TweetTranscriber, BooBCMB, Hillary_Behind_Bars, DuplicateDestroyer, nwordcop, WikipediaPoster, ukpolbot, PaperboyUK, TheHillBot, TrollaBot*

**List of phrases used to remove comments in the Reddit corpora**. "message the mod"; "contact the mod"; "messaging the mod"; "send a message to the mod"; "pm the mod"; "modmail"; "mod mail"; "removed for the following reason"; "removed for breaking rule"; "removed because of the following reason"; "removed as it is a violation of"; "comment has been removed because it violates"; "submission is being removed because"; "submission has been removed as"; "submission has been removed because"; "removed for violating"; "comment has been removed under"; "comment is being removed for"; "comment has been deemed a violation of rule"; "submission has been manually removed by"; "post has been removed due to"; "removed as per rule"; "deleted ^^^^^"; "did not approve this submission, because it doesn't conform to"; "this is an automated message"; "i am a bot"; "i'm a bot"; "i'm a robot"; "i am just a simple bot"; "this message was created by a bot"

**Number of articles used to train the News models**

The number of articles used to trained the News models were obtained from the output of Rozado and al-Gharbi (5)'s analysis scripts (/analysisScripts/countNumberOfArticles - 2015 - 2019.ipynb) that were uploaded with the models.

**Validation of the dimension vectors and the scalar projection method**

We utilized various dictionaries and lexicons related to each dimension to validate the methodology of projecting word vectors onto the dimension vectors. For the morality dimension, we used the Moral Foundation Dictionary 2.0 (6). This dictionary provides words that were deemed virtue or vice words for each moral foundation. We aggregated words across the moral foundations to create one set of virtue words and another set of vice words. These words are referred to as moral and immoral words in the main text. Then, we found words in the common vocabulary set for each set (Reddit: 451 virtue words and 382 vice words; News: 504 virtue words and 411 vice words) and projected the word vectors for these words onto the morality dimension. Finally, we used two-sample *t*-tests to test whether the virtue words have significantly higher scalar projections than the vice words for each model. In all models, virtue words' scalar projections were significantly higher than those of vice words (all *p*s < .001). The distribution of virtue and vice words are visualized with *t* statistics in Figure **S2** (Reddit models) and Figure **S3** (news models).

We used the threat dictionary created by Choi and colleagues (7) to validate the threat dimension. This dictionary includes words that are thought to signal threats. We again found words in the dictionary that appear in the common vocabulary set (Reddit: 197 words; news: 214 words) and projected the word vectors for these words onto the threat dimension. We then tested if the scalar projections of the word vectors of these words were statistically greater than the median scalar projections of all common words via one sample *t*-tests. Across all models, the scalar projections of the threat words from the dictionary onto the threat dimension were significantly higher than the median scalar projections (all *p*s < .001). The distribution of the scalar projections of the threat word vectors onto the threat dimension vector with *t*-statistics are visualized in Figure **S4** (Reddit models) and Figure **S5** (news models).

Finally, we used the NRC Valence, Arousal, and Dominance (NRC-VAD) Lexicon (8), which has aggregated valence rating of individual words (ranging from 0 to 1). For each model, we calculated the correlation (Pearson's *r*) between these ratings and the scalar projections of word vectors for all words that are in the common vocabulary set (Reddit: 7,877 words; News: 9,241 words) onto the valence dimension vector. In all models, the valence rating and the scalar projections were significantly correlated (all *p*s < .001). The correlations are visualized with *r* values in Figure **S6** (Reddit models) and Figure **S7** (news models).

We also note that words used for building the pole vectors for each dimension were not included for analysis concerning that dimension for the validation analysis. This was because these words are projected onto the extreme end due to the setup and might pull the results into false positives. Figure **S1** shows all the test statistics from Figures **S2-7**.

**Difference between our intergroup bias analysis and Rozado and al-Gharbi (5)**

Our approach in the intergroup bias analysis was to build valence, morality, and threat dimension vectors and project word vectors for Democrats-related or Republican-related words onto them. Our results showed that the Republicans-related words projected closer to the pleasant, moral, and safe poles relative to the Democrats-related words in models trained on the conservative text sources, and the opposite pattern occurred in the models trained on liberal text sources. This approach is similar to the approach taken by Rozado and al-Gharbi (5), who analyzed models trained on the news outlets that we used for our analysis of the news outlets. Their approach was to build partisan dimension vectors (Republican to Democrats) and project sentiment lexicons with binary "positive" or "negative" labels on each word to the partisan vector. Then, they calculated the correlation between the semantic labels of the projected words (treating "negative" as -1 and "positive" as 1) and their scalar projections. They found that the correlation calculated this way correlated with the external ideological bias of the news outlets given by Allsides.com. In other words, positive words were closer to the Republican pole compared to negative words in models trained on the conservative outlets and vice versa for the models trained on the liberal outlets. Overall, this analysis is similar to the results from the valence dimension vectors in our approach. However, we stress that we extended their finding by (1) showing that the pattern holds for the morality and threat vectors in the news models and (2) that the pattern holds for models trained on Reddit comments, which Rozado and al-Gharbi did not investigate in their work. Furthermore, we note that our semantic divergence analysis does not overlap with approaches taken by Rozado and al-Gharbi.

**Classification accuracies from "random" dimensions is a more conservative null distribution**

As seen in the manuscript, it is worth noting that comparing against the second null distribution seems to be a more conservative test of classification accuracy. This is expected as the corpus was selected explicitly by partisan bias, and thus there has to be some semantic divergence in general between the corpora. What is surprising is that some "random" dimensions seem to have really high classification accuracy. To further understand this, we qualitatively examined the random dimensions that showed the highest accuracy when sampling one random word for each pole. In this case, it seems that many random dimensions with the highest accuracy have politically polarizing words in at least one pole (e.g., "racial" – "fascism" achieved 97.10% LOOCV accuracy in Reddit and "leftist" – "pentagon" achieved 97.37% LOOCV accuracy in News). This suggests that some random dimensions might arbitrarily capture topics that the partisan are polarized in. However, we also note this is not the case for all high-accuracy random dimensions, and we do not have a definite explanation of why the random dimensions perform above baseline. Table **S7** shows the random dimensions with the highest classification accuracies when using only one randomly sampled word vector for each pole.

**Software Acknowledgment**

The analysis was conducted using programming language Python (https://www.python.org/) version 3.9.12. We are grateful and want to acknowledge for all the developer's and community's efforts that contributed to developing the packages we used for the analyses, including NumPy (9), SciPy (10), pandas (11,12), Gensim (2), Scikit-learn (13), Matplotlib (14), and seaborn (15).

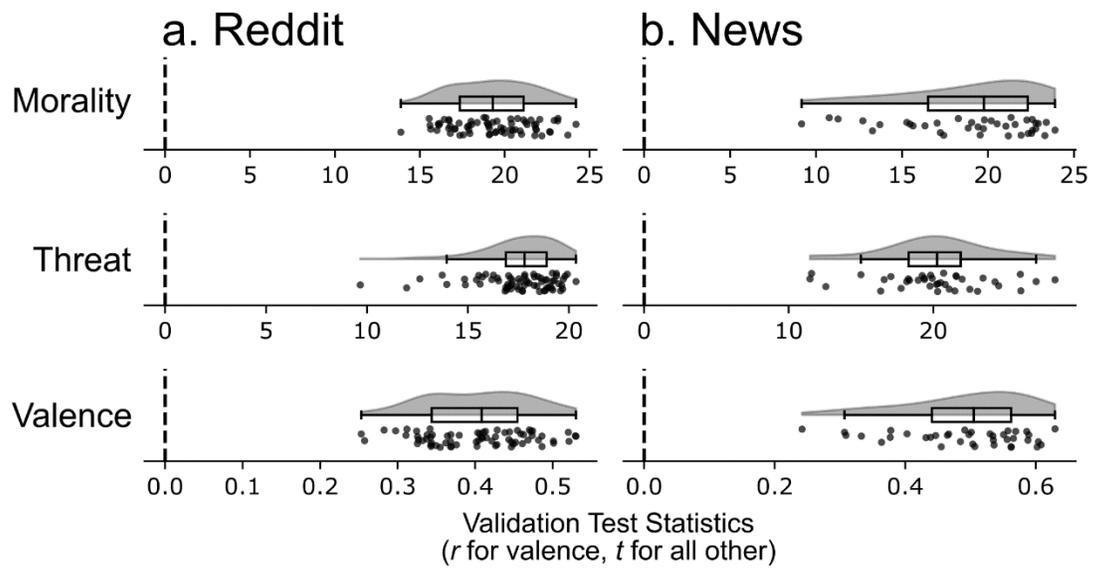

**Figure S1.** Projections of words from external dictionaries onto dimension vectors reflect expected semantic relationships. **a**. Distribution of test statistics for validation tests for each dimension for the Reddit models. Each point denotes the *t* statistic (for morality and threat) or *r* statistic (for valence) for one model. **b**. Distribution of test statistics for validation tests for each dimension for the Reddit models. Plot elements are same with **a**.

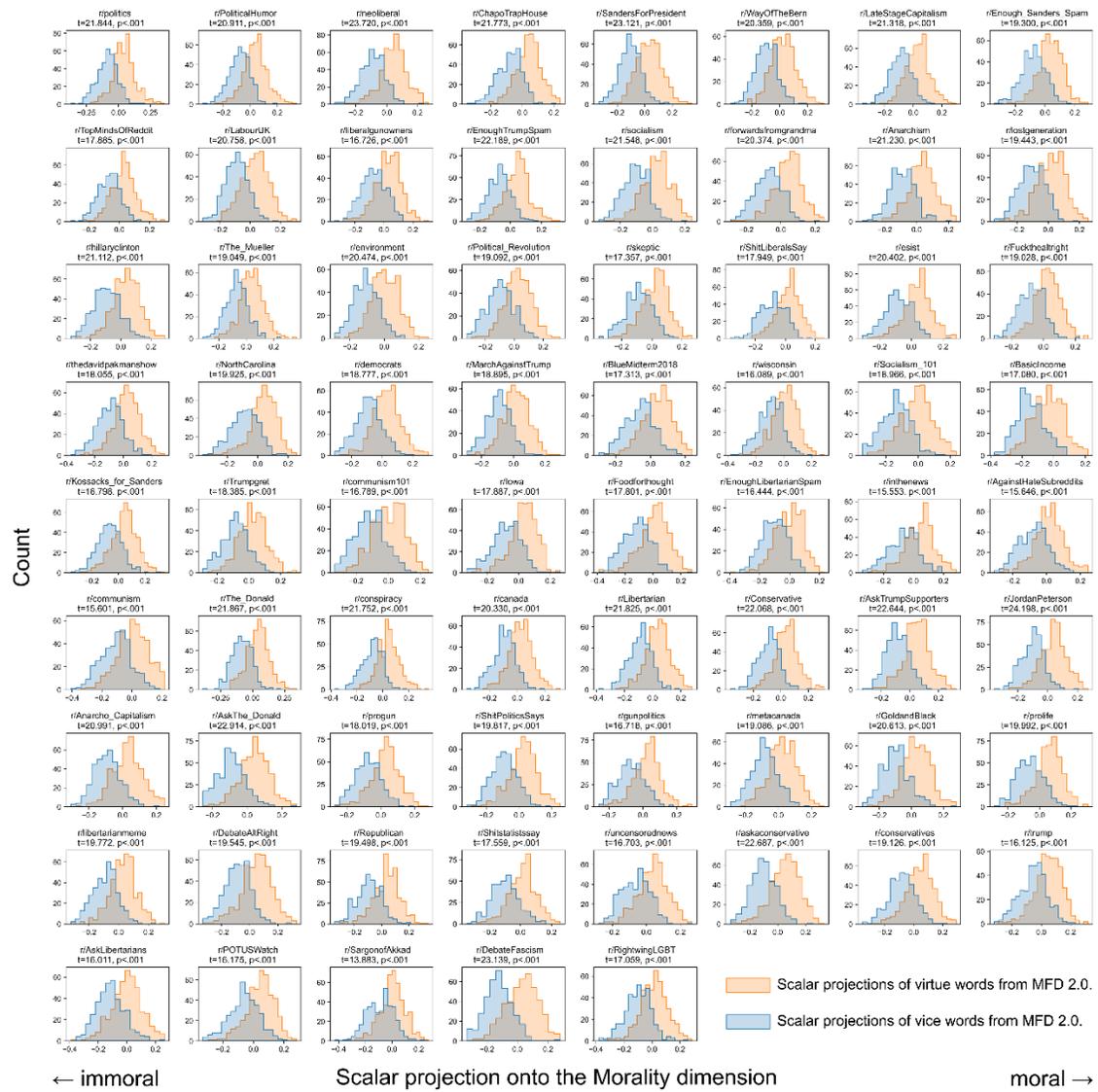

**Figure S2.** The distribution of scalar projections of virtue (orange) and vice (blue) words from the Moral Foundation Dictionary 2.0 is visualized for all Reddit models. The x-axis indicates scalar projections onto the morality dimension, with larger values indicating greater proximity to the moral pole. Distribution of scalar projections of virtue (moral) words are higher than those of vice (immoral) words.

**Figure S3.** The distribution of scalar projections of virtue (orange) and vice (blue) words from the Moral Foundation Dictionary 2.0 is visualized for all news models. The x-axis indicates scalar projections onto the morality dimension, with larger values indicating greater proximity to the moral pole. Distribution of scalar projections of virtue (moral) words are higher than those of vice (immoral) words.

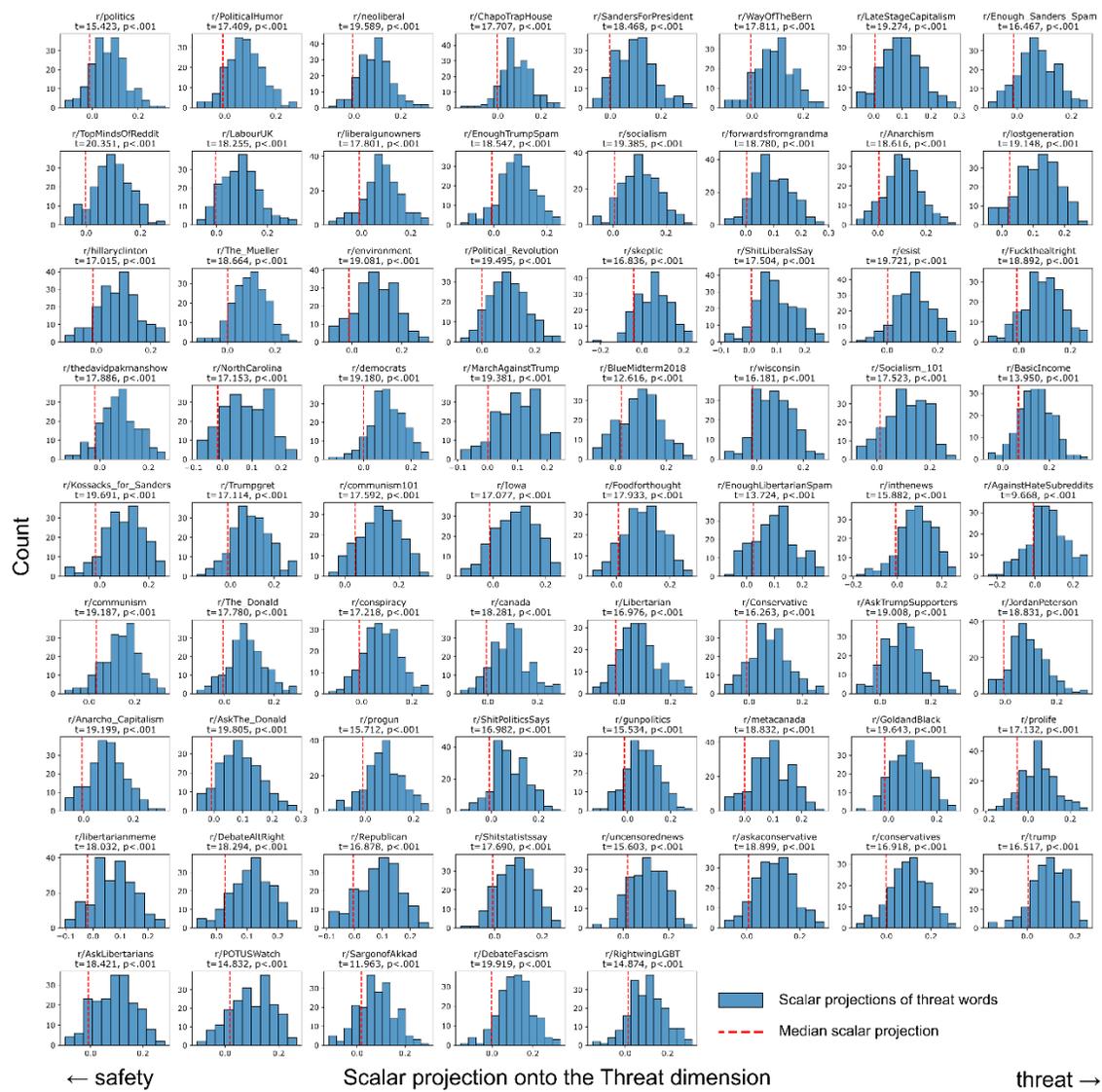

**Figure S4.** The distribution of scalar projections of threat words from the threat dictionary are visualized for all Reddit models. The red line denotes the median scalar projection of all the common words. The x-axis indicates scalar projections onto the threat dimension, with larger values indicating greater proximity to the threat pole. Distribution of scalar projections of threat words are higher than the median scalar projection of all common words.

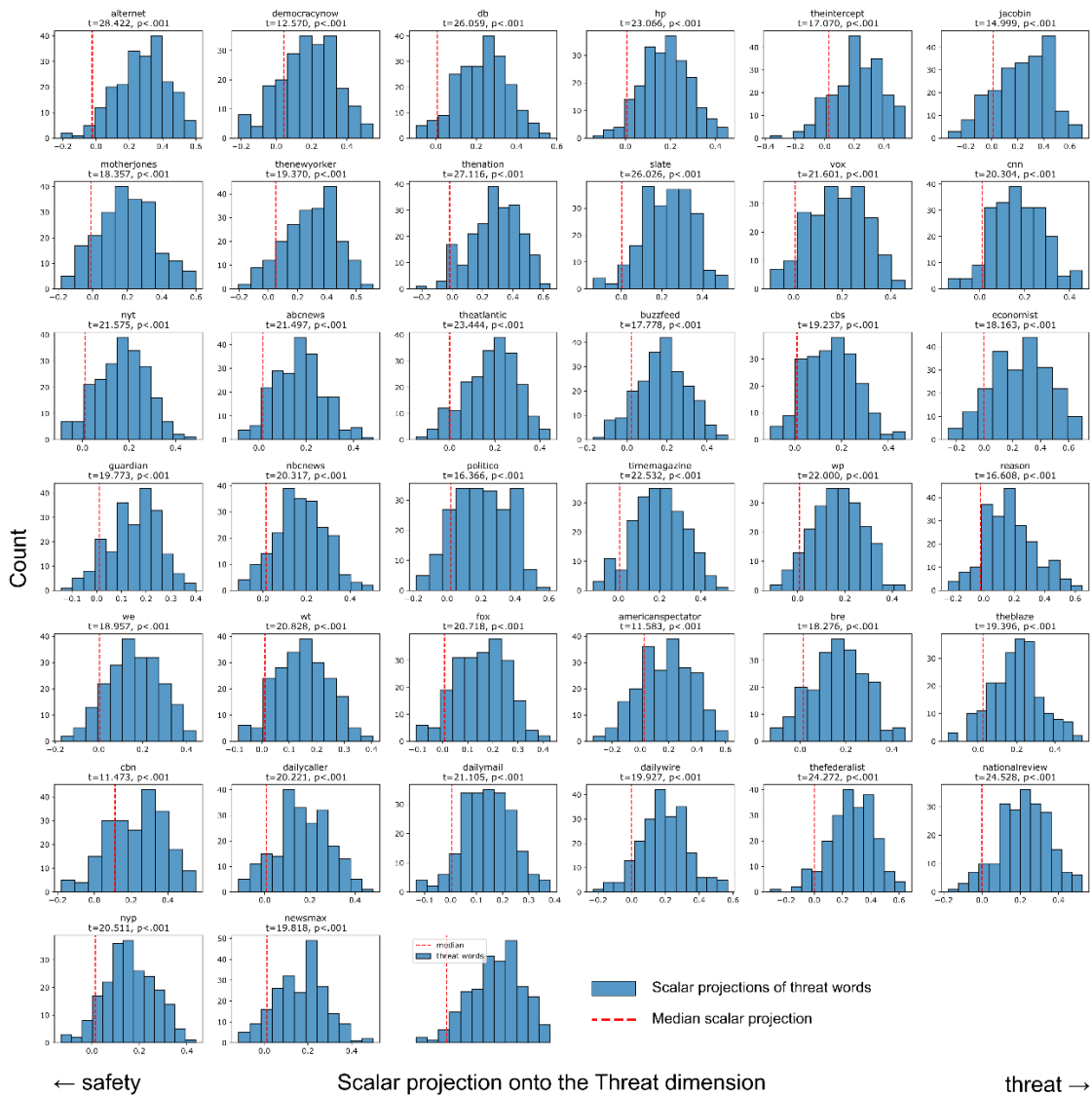

**Figure S5.** The distribution of scalar projections of threat words from the threat dictionary are visualized for all news models. The red line denotes the median scalar projection of all the common words. The x-axis indicates scalar projections onto the threat dimension, with larger values indicating greater proximity to the threat pole. Distribution of scalar projections of threat words are higher than the median scalar projection of all common words.

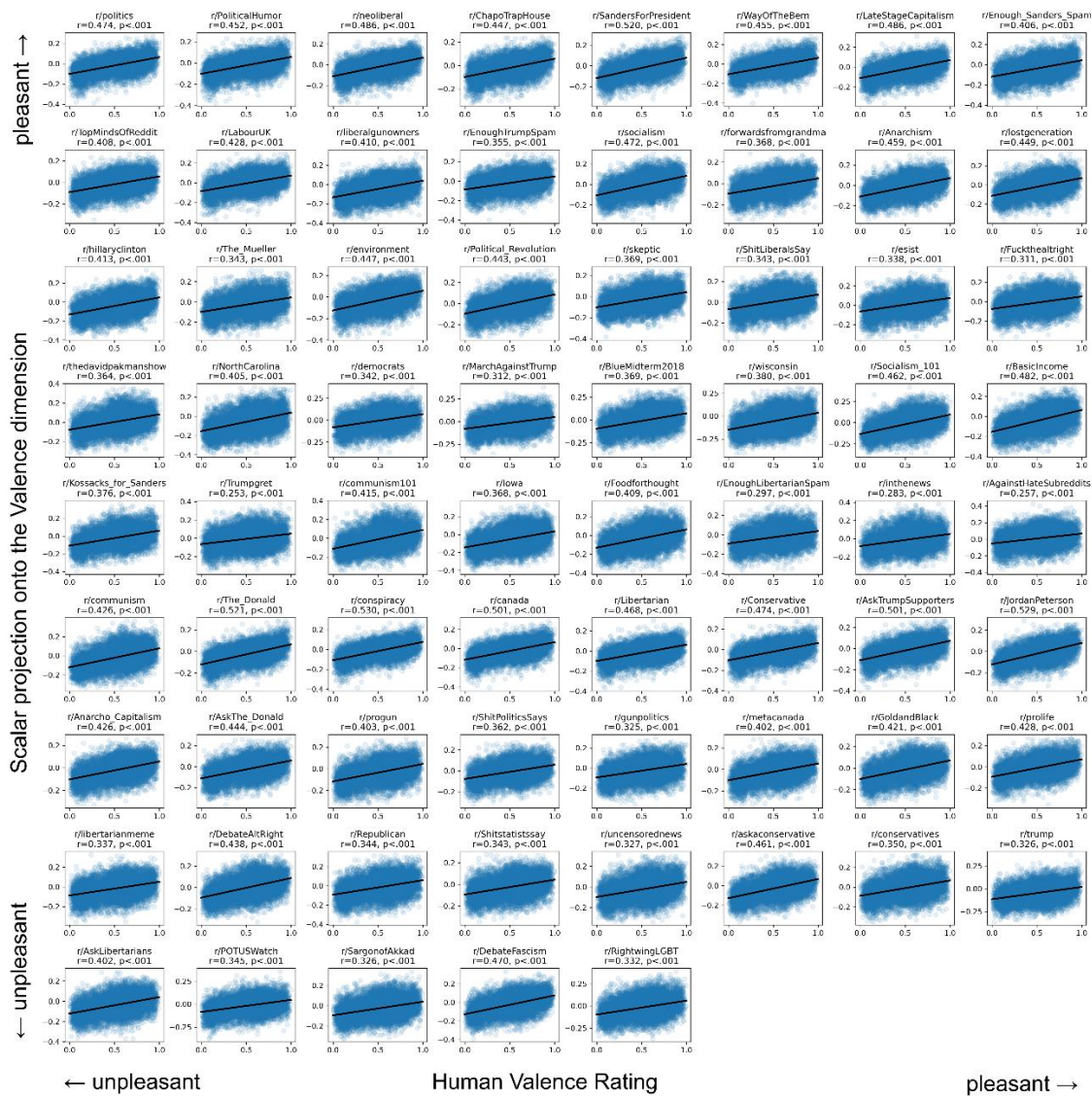

**Figure S6.** The correlation between valence rating from NRC-VAD lexicon (x-axis) and the scalar projections onto the valence dimension vector (y-axis) is visualized for all Reddit models. The black line denotes the best-fit line. The y-axis indicates scalar projections onto the valence dimension, with larger values indicating greater proximity to the pleasant pole. Scalar projections onto the valence dimension are correlated with human valence ratings.

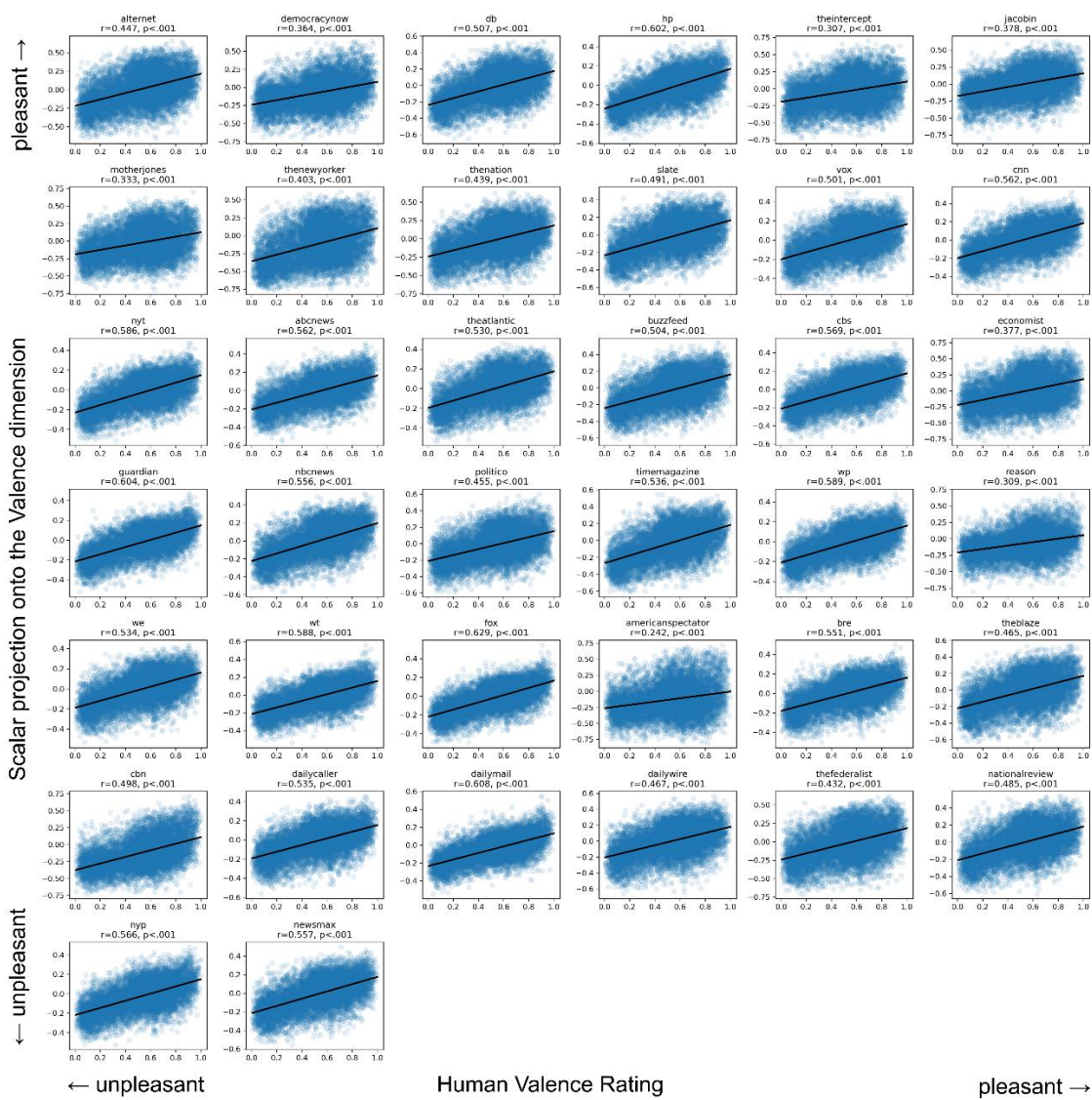

**Figure S7.** The correlation between valence rating from NRC-VAD Lexicon (x-axis) and the scalar projections onto the valence dimension vector (y-axis) is visualized for all news models. The black line denotes the best-fit line. The y-axis indicates scalar projections onto the valence dimension, with larger values indicating greater proximity to the pleasant pole. Scalar projections onto the valence dimension are correlated with human valence ratings.

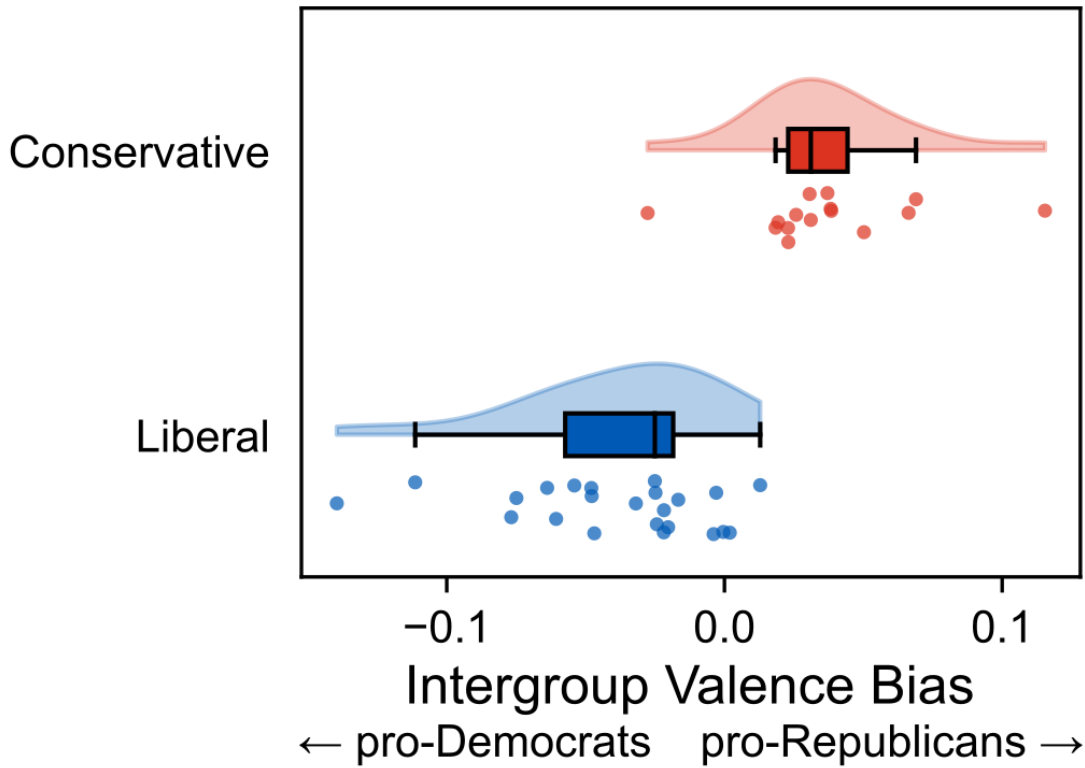

**Figure S8**. Valence association of words related to Democrats and Republicans differed by political bias of the news outlet the models were trained on. Intergroup valence bias refers to the degree to which Republican-related words were closer to the "pleasant" pole than Democrat-related words. The partisan bias was rated by Allsides.com (16). Each point denotes data for one outlet.

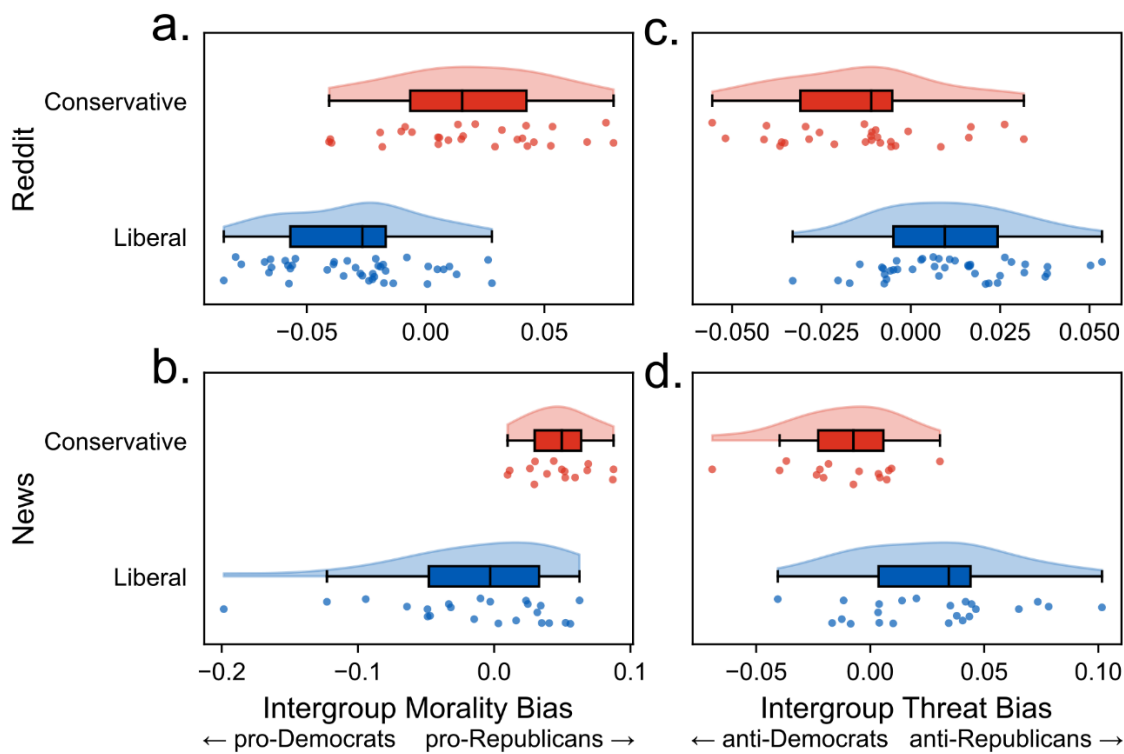

**Figure S9. a**. Morality association of words related to Democrats and Republicans differed by political bias of the subreddit the models were trained on. Intergroup morality bias refers to the degree to which Republican-related words were closer to the "moral" pole than Democrat-related words. The partisan bias was labeled by partisan bias score calculated from user-interaction data by Waller and Anderson **(17)**. **b**. Morality association of words related to Democrats and Republicans differed by political bias of the news outlet the models were trained on. The partisan bias was rated by AllSides.com **(16)**. **c**. Threat association of words related to Democrats and Republicans differed by political bias of the subreddit the models were trained on. Intergroup morality bias refers to the degree to which Republican-related words were closer to the "threat" pole than Democrat-related words. The partisan bias was labeled by partisan bias score calculated from user-interaction data by Waller and Anderson **(17)**. **d**. Threat association of words related to Democrats and Republicans differed by political bias of the news outlet the models were trained on. Each point denotes data for one corpora (subreddit or news outlets) for all subplots.

| Partisanship | Subreddit Name |
|---|---|
| Liberal ($n$=41) | r/AgainstHateSubreddits, r/Anarchism, r/BasicIncome, r/BlueMidterm2018, r/ChapoTrapHouse, r/EnoughLibertarianSpam, r/EnoughTrumpSpam, r/Enough_Sanders_Spam, r/Foodforthought, r/Fuckthealtright, r/Iowa, r/Kossacks_for_Sanders, r/LabourUK, r/LateStageCapitalism, r/MarchAgainstTrump, r/NorthCarolina, r/PoliticalHumor, r/Political_Revolution, r/SandersForPresident, r/ShitLiberalsSay, r/Socialism_101, r/The_Mueller, r/TopMindsOfReddit, r/Trumpgret, r/WayOfTheBern, r/communism, r/communism101, r/democrats, r/environment, r/esist, r/forwardsfromgrandma, r/hillaryclinton, r/inthenews, r/liberalgunowners, r/lostgeneration, r/neoliberal, r/politics, r/skeptic, r/socialism, r/thedavidpakmanshow, r/wisconsin |
| Conservative ($n$=28) | r/Anarcho_Capitalism, r/AskLibertarians, r/AskThe_Donald, r/AskTrumpSupporters, r/Conservative, r/DebateAltRight, r/DebateFascism, r/GoldandBlack, r/JordanPeterson, r/Libertarian, r/POTUSWatch, r/Republican, r/RightwingLGBT, r/SargonofAkkad, r/ShitPoliticsSays, r/Shitstatistssay, r/The_Donald, r/askaconservative, r/canada, r/conservatives, r/conspiracy, r/gunpolitics, r/libertarianmeme, r/metacanada, r/progun, r/prolife, r/trump, r/uncensorednews |

**Table S1.** Subreddits that were analyzed.

| Partisanship | News Outlet |
|---|---|
| Liberal (*n*=23) | ABC News, Alternet, The Atlantic, BuzzFeed News, CBS, CNN, The Daily Beast, Democracy Now!, The Economist, The Guardian, HuffPost, The Intercept, Jacobin, Mother Jones, The Nation, NBC News, The New York Times, The New Yorker, POLITICO, Slate, TIME, Vox, The Washington Post |
| Conservative (*n*=15) | The American Spectator, Breitbart News, TheBlaze, The Christian Broadcasting Network, The Daily Caller, Daily Mail, The Daily Wire, The Federalist, Fox News Channel, National Review, New York Post, NewsMax, Reason, Washington Examiner, The Washington Times |

**Table S2.** News outlets that were analyzed.

| Dimension | Words for Pole 1 | Words for Pole 2 |
|---|---|---|
| Morality | conscience, consciences[N], ethic, ethical, ethically, ethics, fair, fairer, fairly, fairness, goodwill, honest, honestly, honesty, justly[N], moral, moral_superiority[R], morality, morally, morally_superior[R], morals, righteous, righteousness, rightful, rightfully, rightly, virtue, virtues, virtuous | depraved, depravity, evil, evils, foul, guilt, guilty, heinous, immoral, immorality, injustice, injustices, misdeeds, morally_bankrupt[R], nefarious, reprehensible, unconscionable[N], unethical, unfair, unfairly, unfairness, unjust, unjustified, unjustly, vice, vile, wicked, wrong, wrongdoing, wrongful[N], wrongfully, wrongly, wrongs |
| Threat | danger, dangerous, dangerously, dangers, defenseless[R], endanger, endangered, endangering, endangerment[N], endangers, existential_threat[R], harm, harmed, harmful, harming, harms, hazard, hazardous[N], hazards[N], jeopardy, life_threatening[N], peril, perilous[N], perils[N], pernicious[N], risk, risked, risking, risks, risky, threat, threaten, threatened, threatening, threatens, threats, unprotected[N], unsafe | guard, guarded, guardian, guardians, guarding[N], guards, protect, protected, protecting, protection, protections, protective, protector[N], protects, safe, safeguard, safeguarding[N], safeguards, safely, safer, safest, safety, secure, secured, securely[N], securing, security, shelter, sheltered, shelters |
| Valence | amused[R], amusement, amusing[R], awesome, beloved, delight[N], delighted[N], delightful[N], enjoy, enjoyable[R], enjoyed, enjoying, enjoyment, enjoys, fabulous[N], fantastic, fond, fondly[N], glad, gladly, happier, happiest, happily, happiness, happy, joy, joyful[N], joyous[N], magnificent, optimistic, pleasant, please[R], pleased, pleases, pleasing, pleasure, rejoice[N], terrific, uplifting[N], wonderful, wonderfully | abhor[R], abhorrent, abysmal, awful, awfully, depress[N], depressed, depressing, depression, despair, despise, despised, despises[R], discomfort, disheartening, dislike, disliked, dislikes[R], disliking[R], dismal[N], displeasure, distaste[R], distasteful, dreadful[N], gloom[N], grim, heartbreaking, hideous, hopelessly[R], horrendous, horrible, horribly, horrid[R], lament, lamented[N], miserable, miserably, misery, mourn[N], mourning[N], obnoxious, sad, saddened[N], saddest[N], sadly, sadness, terrible, terribly, unfortunate, unfortunately, unhappy, unpleasant, woe[N], woefully, woes, wretched |

**Table S3.** Words used in each pole for calculating the dimension vectors. Note that some words were only used in the Reddit models analysis (denoted with [R]), and some were only used in the News analysis (denoted with [N]). This was due to News and Reddit having different common vocabulary sets.

| Group | Words |
|---|---|
| Republican | conservatism, conservative, conservatives, gop, republican, republican_party[N], republicanism[R], republicans, right_leaning[N], right_wing[N], right_winger[R], right_wingers, rightwing[R] |
| Democrat | dem, democrat, democratic_leader[N], democratic_leaders[N], democratic_party[N], democrats, dems, left_leaning, left_wing, leftie[R], lefties[R], leftism[R], leftist, leftists, leftwing[R], lefty[R], liberal, liberalism, liberals, progressive, progressives, progressivism |

**Table S4.** In-group and out-group words used in the intergroup bias analysis reported in the paper. Note that some words were only used in the Reddit models analysis (denoted with [R]), and some were only used in the News analysis (denoted with [N]). This was due to News and Reddit having different common vocabulary sets.

| Topic | Words |
|---|---|
| Abortion | abortion, abortions, birth, childbirth, fetus, heartbeat, infant_mortality[R], pregnancies[N], pregnancy, pregnant, pro_choice[R], reproductive, trimester[N] |
| Constitution | amendment, amendments, constitution, constitutional, constitutionality[N], constitutionally, constitutions, unconstitutional, unconstitutionally[N] |
| Guns | firearm, firearms, gun, gun_control[R], gunfire[N], gunman[N], gunmen[N], gunned[N], gunning[N], gunpoint, guns, gunshot[N], handgun, handguns, homicide, homicides[N], mass_shootings[R], murder, murdered, shoot, shooter, shooters, shooting, shootings, shoots, shotgun, shotguns[N] |
| Immigration | border, borders, caravan[N], caravans[N], hispanic, hispanics, illegal, illegal_immigrants[R], illegals[N], immigrant, immigrants, immigrate, immigrated, immigration, mexican, mexicans, mexico, migrant, migrants, migrate, migrated, migrating[N], migration, open_borders[R], xenophobia, xenophobic |
| LGBTQ+ Community | bisexual[N], gay, gay_marriage[R], gays, homo[R], homophobe[R], homophobia, homosexual, homosexuality, homosexuals, lesbian, lesbians[N], queer, sex_marriage[R], transgender, transgendered[R] |
| Police and Criminals | committing_crimes[R], crime, crimes, criminal, criminal_activity[R], criminality, criminalize[R], criminalized[R], criminalizing[R], criminally, criminals, decriminalization[R], jail, jails, police, police_brutality[R], police_departments[R], police_officer, police_officers[R], policeman, policemen[N], polices[R] |
| Religion | god, gods, religion, religions, religiosity[N], religious, religious_beliefs[R], religiously |

**Table S5.** Selected words used in the semantic divergence analysis using issue words reported in the paper by topics. Note that some words were only used in the Reddit models analysis (denoted with [R]), and some were only used in the News analysis (denoted with [N]). This was due to News and Reddit having different common vocabulary sets.

| Dimension | Words for Pole 1 | Words for Pole 2 |
|---|---|---|
| Size | big, bigger, biggest, colossal, cosmic[N], enormous, enormously, giant, giants, gigantic, huge, hugely, immense, immensely, large, largely, larger, largest, massive, massively | compact[N], dwarf[R], little, micro, mini, minimal, minimally[N], minimize[R], minimized[R], minimizing[R], minuscule[R], portable[N], small, small_fraction[R], small_percentage[R], smaller, smallest, tiny |
| Speed | fast, faster, fastest, haste[N], hasten[N], hastily, promptly, quick, quicker, quickest[R], quickly, rapid, rapidly, speedy[N], swiftly | crawling, gradual, gradually, lag[N], slow, slowdown[N], slowed[N], slowed_down[R], slower, slowing, slowly, slows[N] |
| Wetness | drenched[N], rain, rainfall[N], raining, rains[N], rainy[N], soaked, swim, swimming, water, wet | desert, deserts, dried, drought[N], droughts[N], dry, drying[N], sun, sunny[N] |

**Table S6.** Words used in each pole for calculating the dimension vectors for the theoretically irrelevant dimensions. Note that some words were only used in the Reddit models analysis (denoted with [R]), and some were only used in the News analysis (denoted with [N]). This was due to News and Reddit having different common vocabulary sets.

| Corpora | Classification Accuracy | Random Dimension Poles |
|---|---|---|
| Reddit | 97.1% (67 out of 69) | ('racial', 'fascism'), ('minority', 'academia') |
| Reddit | 95.65% (66 out of 69) | ('leftism', 'vodka'), ('fascism', 'flaw'), ('picking', 'privilege'), ('clarification', 'minimum_wage'), ('quality', 'cruelty'), ('myself', 'spectrum'), ('laws', 'both_sides') |
| Reddit | 94.2% (65 out of 69) | ('white_supremacist', 'expertise'), ('immigrant', 'hat'), ('awful', 'neo_nazi'), ('white_supremacists', 'abusing'), ('fascist', 'fits'), ('conservatism', 'deflect'), ('unsubstantiated', 'wealthy'), ('minorities', 'ridiculous'), ('supremacy', 'degrees'), ('sovereign', 'white_supremacy'), ('gop', 'decay'), ('fascists', 'flooding'), ('oppressors', 'leftist'), ('academia', 'gambling'), ('bigot', 'influences'), ('dude', 'loves'), ('satisfied', 'intellectuals'), ('calling', 'deal'), ('emperor', 'privilege'), ('privileged', 'spectrum'), ('very', 'dealt'), ('common_sense', 'wing'), ('naive', 'investigating'), ('guilt', 'anxiety'), ('arrested', 'drives'), ('visiting', 'sycophants'), ('history', 'youth'), ('entrepreneur', 'refusing') |
| News | 97.37% (37 out of 38) | ('leftist', 'pentagon'), ('baltic', 'teaching') |
| News | 94.74% (36 out of 38) | ('democrats', 'conservative'), ('theocracy', 'summary'), ('hourly', 'smash'), ('holders', 'nailed') |
| News | 92.11% (35 out of 38) | ('socialist', 'trans'), ('christianity', 'stories'), ('aircraft', 'supremacy'), ('guilt', 'missionary'), ('colonialism', 'followers'), ('explicitly', 'rock'), ('understood', 'post'), ('pleas', 'businesses'), ('averaged', 'christine'), ('consequence', 'matching'), ('secular', 'crying'), ('emerald', 'enforced'), ('toni', 'lows'), ('motorcycle', 'divest'), ('kept', 'sense'), ('allocate', 'parody'), ('slipped', 'wreak'), ('minds', 'romney'), ('wrestled', 'leftist'), ('establishment', 'enterprise'), ('administrative', 'patron'), ('person', 'stalin'), ('masculinity', 'magic'), ('counties', 'teacher'), ('ignore', 'imagine'), ('plotting', 'importers') |

**Table S7**. Random dimensions with best LOOCV accuracy when using one randomly sampled word for each pole.